\documentclass[11pt]{latex/aime}

\usepackage[T1]{fontenc}

\AtBeginDocument{\urlstyle{same}}

\usepackage{booktabs}
\usepackage{array}
\usepackage{xcolor}
\usepackage{svg}

\usepackage{multirow}
\usepackage{adjustbox}
\usepackage{longtable}
\Crefname{appendix}{Appendix}{Appendices}
\crefname{appendix}{appendix}{appendices}

\title{%
\parbox{\textwidth}{%
    \raggedright
    \normalfont\rmfamily\bfseries\scshape
    \fontsize{16.5}{18}\selectfont
    TACFormer: Transformer-Based Generative Model\\[-2pt]
    for Target-Conditioned \textit{De Novo} PROTAC Design%
}%
\par\vspace{10pt}%
}

\author[1]{Yossra Gharbi}
\author[1,2,*]{Rocío Mercado}

\affiliation[1]{%
AI Laboratory for Molecular Engineering (AIME),
Department of Computer Science and Engineering,
Chalmers University of Technology \&
University of Gothenburg, Gothenburg, Sweden
}
\affiliation[2]{%
Science for Life Laboratory (SciLifeLab), Gothenburg, Sweden
}

\contribution[*]{Corresponding author}

\correspondence{%
\email{rocio.mercado@chalmers.se}%
}

\codeavailability{%
\url{https://github.com/yossragh/TACFormer}%
}

\date{\today}

\abstract{%
Targeted protein degradation (TPD) represents a fundamental shift in drug discovery, moving beyond traditional target inhibition toward the event-driven removal of disease-causing proteins.
A prominent strategy for TPD is the use of proteolysis-targeting chimeras (PROTACs), heterobifunctional molecules that selectively route a protein of interest (POI) to the ubiquitin--proteasome system for degradation. While PROTAC design still relies heavily on trial-and-error synthesis and experimental testing, deep generative models offer a promising means to explore PROTAC chemical space \textit{in silico} and generate novel molecules.
However, many existing generative methods formulate PROTAC design primarily as a linker-generation problem, with POI-binding (warhead) and E3-ligase-binding components fixed in advance. This restricts exploration of the full PROTAC chemical space and often leaves the POI implicit through the predefined warhead, rather than using protein-level information as an explicit conditioning signal during generation.
To address these limitations, we propose TACFormer: a transformer-based generative model for target-aware \textit{de novo} PROTAC design. The model conditions a GPT-style molecular decoder on pretrained ESM-2 representations of POI sequences, allowing POI-level information to steer full-molecule generation.
Our results show that TACFormer generates highly novel and chemically diverse PROTACs while showing consistent POI-dependent shifts in generated chemical space. In particular, molecules generated under specific target conditions preferentially populate chemical neighborhoods associated with POIs in the held-out dataset, while extrapolating to novel chemotypes. By natively integrating biological context into the generative process, TACFormer moves PROTAC design toward a more biology-informed exploration of degrader chemical space.
}

\begin{document}

\maketitle

% ================================================================
% INTRODUCTION
% ================================================================

% Your manuscript sections can be inserted here.

% \section{Introduction}
% \section{Related Work}
% \section{Methods}
% \section{Results}
% \section{Discussion}
% \section{Conclusion}

% ================================================================
% AUTHOR CONTRIBUTIONS
% ================================================================

\section{Introduction}
Targeted protein degradation (TPD) has emerged as a promising paradigm in drug discovery, shifting from sustained occupancy of a functional site to the event-driven removal of the target protein itself~\citep{bekes2022protac}.
Because TPD does not require a druggable active site, it can extend to targets that have long been considered ``undruggable''~\citep{hopkins2002druggable,bushweller2019targeting}. Proteolysis-targeting chimeras (PROTACs) are among the most developed TPD modalities, with several having advanced into clinical trials~\citep{o2023delivering}. A PROTAC is a heterobifunctional molecule composed of a warhead that binds a protein of interest (POI), an E3 ligand that recruits an E3 ubiquitin ligase, and a linker that tethers the two components~\citep{zhao2022targeted}. By simultaneously binding the POI and E3 ligase, a PROTAC brings the two proteins into proximity, inducing polyubiquitination of the POI and its subsequent degradation by the proteasome~\citep{zhao2022targeted}.

Despite this promise, PROTAC discovery still relies heavily on iterative synthesis and experimental screening, which must be repeated for each new POI~\citep{pike2026lessons}. Deep generative models offer an alternative by learning from known chemistry and proposing molecules beyond existing libraries~\citep{pang2024deep}. Although three-dimensional molecular generation is gaining attention, language-based models remain among the most sample-efficient approaches, and string-based molecular representations continue to be widely used~\citep{skinnider2021chemical,grisoni2023chemical,guo2026sample}.

These models typically represent molecules using the simplified molecular-input line-entry system (SMILES). SMILES is non-injective, meaning that the same molecule can be represented by multiple valid strings because atoms can be traversed in different orders~\citep{weininger1989smiles}. This property offers two advantages for molecular generation. First, randomized SMILES can improve the quality and coverage of generated chemical space~\citep{arus2019randomized}. Second, different representations of the same molecule provide a natural form of data augmentation, helping models learn from scarce  datasets~\citep{skinnider2021chemical,moret2020generative}. This is particularly valuable for PROTAC generation, where only a few thousand experimentally characterized degraders are available.

Existing generative methods for PROTAC design largely focus on linker generation~\citep{gharbi2024comprehensive}, requiring the warhead and E3 ligand to be known beforehand. This confines the search to the linker and restricts \textit{de novo} design to targets for which suitable binders are already known. As a result, these methods cannot jointly optimize the complete PROTAC, narrowing the explorable chemical space, notably by excluding the discovery of novel warheads. 

The few methods that generate complete degraders introduce target information through intermediate proxies, such as target degradation predictors~\citep{nori2022protac}, affinity-based fragment selection~\citep{shao2024lmprotac}, docking-based rewards~\citep{xu2026synprotac}, or downstream optimization objectives~\citep{dai2024trlprotac}, rather than conditioning the molecular generator on the target protein itself. This is an important limitation because the chemical structure of an effective PROTAC fundamentally depends on the biological properties of the POI it is intended to degrade~\citep{tsai2024targeted}. To our knowledge, no existing method uses a learned representation of the target protein to condition \textit{de novo} generation of complete PROTAC molecules.

In this work, we aim to address the gaps in existing work by developing a method that (i) generates complete PROTAC molecules rather than restricting generation to the linker; (ii) integrates the biological representation of the POI as a conditioning signal; and (iii) operates effectively in low-data regimes inherent to PROTAC discovery. We summarize our contributions as follows:
\begin{itemize}
    \item We introduce TACFormer, which is, to our knowledge, the first generative model that directly conditions \textit{de novo} full-molecule PROTAC generation on a learned representation of the target protein. We use pretrained ESM-2~\citep{lin2023evolutionary} to encode the POI amino-acid sequence and aggregate the resulting residue embeddings into a fixed-length protein representation. This representation conditions a GPT-style decoder-only transformer that generates PROTAC molecules as SMILES.

    \item We address the low-data regime through two-stage training. We curate 79,406 PROTAC-like molecules by filtering ZINC20 and ChEMBL 36 for physicochemical and structural properties characteristic of degraders, and use this corpus to pretrain a PROTAC-oriented molecular prior. We then fine-tune the pretrained model on reported PROTACs. We also contribute a normalization pipeline that harmonizes POI and E3-ligase annotations across the three PROTAC source databases.

    \item We establish that POI identity carries a measurable chemical signature in PROTAC data. Molecules targeting the same POI can span multiple chemotypes, but their local chemical neighborhoods remain strongly enriched for that target. This reveals that POI-associated chemistry is globally heterogeneous yet locally strongly organized.

    \item We propose an evaluation protocol for target-conditioned molecular generation that combines a likelihood-based measure ($\Delta_{\mathrm{NLL}}$) with nearest-neighbor purity in chemical space. We use these complementary metrics to assess both where conditioning succeeds and where it fails, including zero-shot evaluation on POIs unseen during training.
\end{itemize}

\begin{figure}[t]
    \centering
    \includegraphics[width=\linewidth]{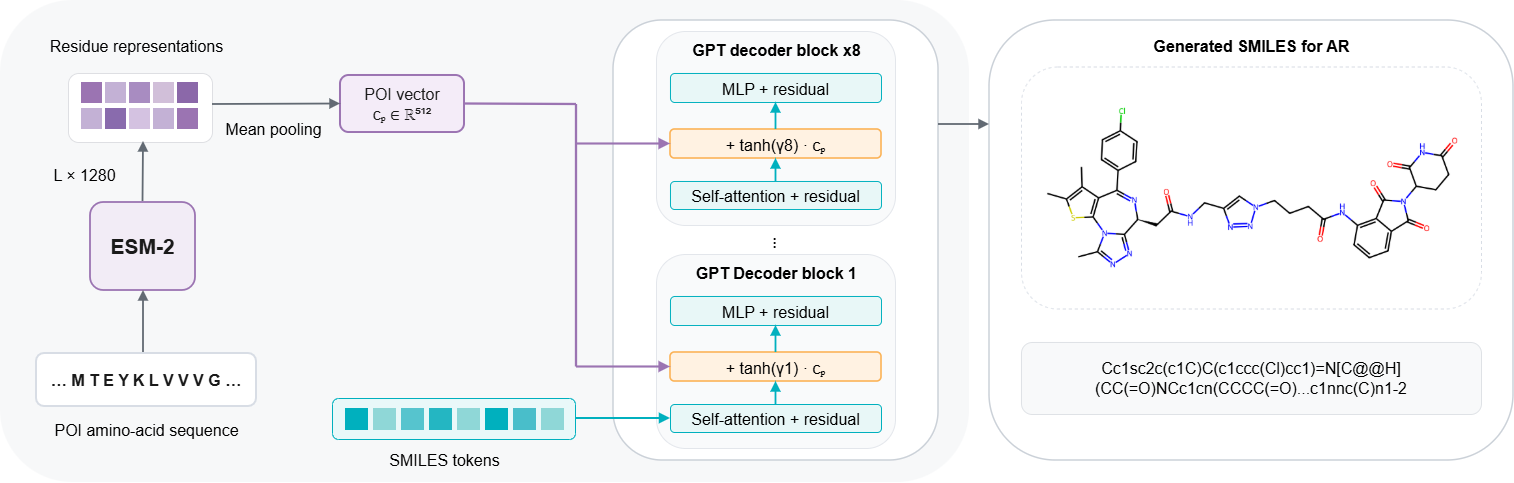}
    \caption{\textbf{Overview of the TACFormer architecture.}  The protein of interest (POI) amino-acid sequence is encoded using pretrained ESM-2. The resulting residue representations are mean-pooled and projected into a 512-dimensional POI embedding. A GPT-style decoder generates PROTAC SMILES autoregressively. Protein conditioning is introduced in each decoder block by adding the projected POI embedding to the hidden representations after self-attention, scaled by a learnable layer-specific gate. The model generates a complete PROTAC SMILES conditioned on the specified POI.}
    \label{fig:tacformer_architecture}
\end{figure}

\section{Related work}
\paragraph{Transformer-based molecular generation.}
Transformer architectures are widely used for molecular design with string-based representations~\citep{grisoni2023chemical}. Encoder--decoder models can handle sequence-to-sequence tasks such as reaction prediction~\citep{schwaller2019molecular,tetko2020state}, retrosynthesis~\citep{schwaller2020predicting,lin2020automatic,irwin2022chemformer}, and molecular optimization~\citep{irwin2022chemformer,he2022transformer}. Encoder-only models learn bidirectional representations of molecular sequences and are mainly used for predictive tasks, including molecular property prediction, activity modeling, and compound--protein interaction prediction~\citep{yu2021review}. Decoder-only models use causal self-attention to generate molecular sequences autoregressively, making them naturally suited to molecular generation conditioned on physicochemical properties~\citep{bagal2021molgpt,lu2023graphgpt}, scaffolds~\citep{bagal2021molgpt}, or target-related information~\citep{wang2023cmolgpt}. Conditioning on protein sequence information is less common: Chen \& Bajorath\citep{chen2024generative} use protein-language-model embeddings to condition a SMILES decoder and generate compounds with desired potency against a target protein, but consider single-ligand rather than bifunctional design. Across these architectures, pretraining on a broad molecular corpus followed by fine-tuning on smaller domain-specific datasets supports transfer learning to data-limited settings~\citep{grisoni2023chemical,mswahili2024transformer}.

\paragraph{Generative PROTAC linker design.}
Most generative work on PROTACs formulates the problem as linker design~\citep{gharbi2024comprehensive}. Early methods apply general-purpose fragment-linking models trained on small-molecule data to PROTAC linker design, without training on actual PROTAC data~\citep{imrie2020deep,guo2023link}. Subsequent methods are tailored to PROTAC linker generation through PROTAC-specific training, fine-tuning, or optimization objectives that account for their chemical and structural requirements. These include sequence-based autoregressive generation optimized with reinforcement learning~\citep{zheng2022accelerated}, graph-based linker generation~\citep{kao2023fragment}, and approaches incorporating three-dimensional objectives~\citep{neeser2023reinforcement} or diffusion-based modeling~\citep{li2024diffprotacs,song2025domain}. All these methods require the warhead and E3 ligand as input, restricting generation to the linker and leaving the combinatorial space of the whole molecule unexplored.

\paragraph{Full-molecule PROTAC generation and target information.}
Few methods address full-molecule PROTAC generation. Nori et al.\citep{nori2022protac} generate PROTAC-like molecular graphs from an empty graph and introduce target specificity through reinforcement learning, using a surrogate model that predicts degradation activity for a given target as the reward. TRL-PROTAC~\citep{dai2024trlprotac} generates warhead and E3 ligands separately, connects them using a linker model, and optimizes the assembled PROTACs using pharmacokinetic and degradation scores. Shao et al.\citep{shao2024lmprotac} use a protein--compound affinity model to select target-relevant fragments, which guide PROTAC generation under structural and physicochemical constraints. SynPROTAC~\citep{xu2026synprotac} assembles synthesizable PROTACs from a supplied warhead or E3 ligand using reaction-based molecular assembly and uses docking-guided reinforcement learning to optimize binding-related properties. Although these methods are not target-agnostic, they do not directly condition PROTAC generation on representations of the target protein.

\section{Preliminaries}
\label{sec:preliminaries}

\subsection{Autoregressive formulation}

PROTACs are represented as SMILES strings, allowing molecular generation to be formulated as autoregressive language modeling. For a molecule tokenized as \begin{equation}
\mathbf{s}=(s_1,\ldots,s_T),
\end{equation}
the conditional distribution is factorized as
\begin{equation}
P_{\theta}(\mathbf{s}\mid \mathbf{z}_p)
=
\prod_{t=1}^{T}
P_{\theta}(s_t\mid s_{<t},\mathbf{z}_p).
\end{equation}
where $\theta$ denotes the model parameters, $p$ the conditioning POI, and $\mathbf{z}_p$ its fixed protein embedding, defined below. The conditional model therefore learns a family of POI-dependent molecular distributions rather than a single unconditional distribution over PROTACs. Tokenization and SMILES randomization are described in Appendix~\ref{app:tokenization}.

\subsection{Protein representation using ESM-2}

We represent each POI using ESM-2~\citep{lin2023evolutionary}, a pretrained transformer protein language model that learns contextual residue representations from amino-acid sequences.
These representations capture sequence patterns related to protein evolution and structure.
For each POI, we retrieve its amino-acid sequence from UniProtKB~\citep{ahmad2025uniprot} and encode it using the pretrained \texttt{esm2\_t33\_650M\_UR50D} variant (33 layers, 650M parameters, hidden size 1280).
For a protein of length $L$, the resulting residue representations are
\begin{equation}
\mathbf{H}_p
=
[\mathbf{h}_1,\ldots,\mathbf{h}_L]^{\top}
\in
\mathbb{R}^{L\times1280}
\end{equation}
We apply mean pooling over the residue representations to obtain a fixed-length POI embedding,
\begin{equation}
\mathbf{z}_p
=
L^{-1}
\sum_{i=1}^{L}
\mathbf{h}_i
\in
\mathbb{R}^{1280}.
\end{equation}
These sequence-derived embeddings allow molecular generation to be conditioned on protein sequence information rather than discrete target identifiers or static protein descriptors. 

Proteins longer than the 1022-residue processing window are encoded in overlapping windows.
The residue representations are merged to reconstruct the full-length sequence before mean pooling, ensuring that each residue contributes once to the final embedding (Appendix~\ref{app:long_sequences}).
ESM-2 remains frozen throughout training.

\subsection{Conditional decoder}

The molecular generator is a decoder-only transformer with eight pre-LayerNorm blocks, a hidden dimension of $d_{\mathrm{model}}=512$, and eight attention heads (Appendix~\ref{app:decoder_details}). 
The same backbone is used for molecular pretraining, unconditional fine-tuning, and target-conditioned fine-tuning.

The pooled protein representation is projected into the decoder hidden space as
\begin{equation}
\mathbf{c}_p
=
\mathbf{W}_c
\operatorname{LN}_p(\mathbf{z}_p)
+
\mathbf{b}_c,
\qquad
\mathbf{W}_c
\in
\mathbb{R}^{512\times1280}.
\end{equation}

We introduce protein conditioning through gated residual addition in each transformer block.
The projected protein embedding $\mathbf{c}_p$ is broadcast across token positions and added after the causal self-attention residual, scaled by a trainable layer-specific gate $\gamma_\ell$.
For block $\ell$, the attention residual is
\begin{equation}
\mathbf{A}^{(\ell)}
=
\mathbf{X}^{(\ell)}
+
\operatorname{MHA}
\left(
\operatorname{LN}_1
\left(
\mathbf{X}^{(\ell)}
\right)
\right),
\end{equation}
and the conditioned block update is
\begin{equation}
\widetilde{\mathbf{A}}^{(\ell)}
=
\mathbf{A}^{(\ell)}
+
\tanh(\gamma_{\ell})\,\mathbf{c}_p,
\qquad
\mathbf{X}^{(\ell+1)}
=
\widetilde{\mathbf{A}}^{(\ell)}
+
\operatorname{MLP}
\left(
\operatorname{LN}_2
\left(
\widetilde{\mathbf{A}}^{(\ell)}
\right)
\right).
\end{equation}

The protein condition $\mathbf{c}_p$ is shared across all transformer blocks, while the layer-specific gates $\gamma_\ell$ control its contribution at each depth.
The conditional model is initialized from the pretrained molecular prior and augmented with the protein adapter and one gate per block.
Each gate is initialized to $\gamma_\ell=0$, so $\tanh(\gamma_\ell)\mathbf{c}_p=\mathbf{0}$ and the model initially behaves identically to the pretrained molecular decoder.
Protein conditioning is then learned during fine-tuning using teacher-forced causal language modeling on molecule--POI pairs.

\section{Experiments}
\label{sec:experiments}

\subsection{Data}
\label{sec:data}

\paragraph{PROTAC-like pretraining data.}
To learn a PROTAC-oriented molecular prior, compounds from ZINC20~\citep{irwin2020zinc20} and ChEMBL 36~\citep{zdrazil2024chembl} are filtered for physicochemical and structural properties characteristic of PROTAC-like molecules. After standardization, filtering, and deduplication, the pretraining corpus contains 79,406 unique molecules. The filtering criteria are detailed in Appendix~\ref{app:pretraining_corpus}.

\paragraph{PROTAC data.}
Two datasets are built by merging PROTAC-DB~\citep{ge2025protac}, TPDdb~\citep{qin2026tpddb}, and PROTAC-Pedia~\citep{protacpedia}.

For conditional generation, each observation is a molecule--POI pair. We harmonize target annotations across the three databases by mapping POI names to UniProt accessions and using the corresponding primary gene names as canonical labels. After deduplication, the dataset contains 30,053 molecule--POI pairs across 408 POIs.   

We reserve four POIs---WDR5, DHFR, FGFR3, and MAP3K13---for zero-shot evaluation. These targets span different levels of similarity to the training POIs in ESM-2 embedding space. 
We verify that molecules targeting the zero-shot POIs do not occur under any training POI.
Removing their 217 molecule--POI pairs before splitting leaves 29,836 pairs across 404 POIs.

Because the data are unevenly distributed across targets, we select the 20 most represented POIs for target-wise held-out evaluation, hereafter referred to as the \emph{seen POIs}. These targets account for 21,246 pairs (71.2\% of the remaining dataset), providing sufficient reference molecules for separate evaluation of each POI.

For the seen POIs, we split pairs into training (80\%), validation (10\%), and test (10\%) subsets. The remaining 384 POIs contribute 8,590 pairs exclusively to training, exposing the model to a broader range of protein conditions. To prevent molecular leakage, all pairs sharing the same standardized SMILES are assigned to one partition. The final conditional dataset contains 25,584 training, 2,126 validation, and 2,126 test pairs.

For unconditional generation, we discard POI annotations from the conditional partitions and deduplicate standardized SMILES within each partition. This removes repeated molecules with multiple POI labels while preserving their split assignments. The resulting dataset contains 22,880 (84.37\%) training, 2,122 (7.82\%) validation, and 2,117 (7.81\%) test molecules. Both models therefore use matched molecular partitions.

Dataset composition, POI harmonization, zero-shot target selection, and splitting procedures are provided in Appendices~\ref{app:dataset_composition}, \ref{app:poi_normalization}, \ref{app:zero_shot_selection}, and~\ref{app:data_splits}.

\subsection{Experimental setup}
\label{sec:experimental_setup}

Both downstream models are initialized from the same molecular prior. We first pretrain the decoder-only transformer on PROTAC-like molecules, then fine-tune separate unconditional and conditional models on PROTAC data.

The \emph{unconditional model} is fine-tuned without protein information, learning a distribution $P_{\phi}(\mathbf{s})$ over complete PROTAC molecules. The \emph{conditional model} is fine-tuned on molecule--POI pairs using the conditioning mechanism described in Section~\ref{sec:preliminaries}, learning $P_{\theta}(\mathbf{s}\mid\mathbf{z}_p)$.

Both models share the same molecular backbone. For each sampling seed, the unconditional model generates 200,000 molecules. The conditional model generates 10,000 molecules per POI, yielding the same total generation budget.

We evaluate the conditional generation in two settings. In the \emph{seen-POI setting}, the model generates molecules for the seen POIs and is evaluated against their held-out test data. In the \emph{zero-shot setting}, the same model generates molecules for WDR5, DHFR, FGFR3, and MAP3K13, which are excluded from training, without additional fine-tuning. Each zero-shot target receives 10,000 generated samples per seed.

Training hyperparameters and sampling settings are provided in Appendices~\ref{app:optimization_details} and~\ref{app:evaluation_details}, respectively.

\subsection{Evaluation}
\label{sec:evaluation}

Let $\mathcal{G}$ denote the set of generated SMILES, $\mathcal{V}\subseteq\mathcal{G}$ the subset of samples successfully parsed and sanitized by RDKit, and $\mathcal{U}$ the set of unique valid molecules obtained by canonicalizing and deduplicating $\mathcal{V}$. For conditional generation, we define the corresponding collections $\mathcal{G}_p$, $\mathcal{V}_p$, and $\mathcal{U}_p$ for each POI $p$.  Metrics are computed per POI and summarized by an unweighted macro average across targets. Implementation details are provided in Appendix~\ref{app:evaluation_details}.

\paragraph{Evaluation rationale.}
We first examine whether PROTAC chemistry in the reference data contains a POI-specific signal that can support conditional generation. We compare Morgan--Tanimoto similarity between molecules targeting the same POI and those targeting different POIs. Because many PROTACs share E3-binding ligands, whole-molecule similarity may partly reflect shared E3 identity rather than the POI. We therefore repeat this comparison within CRBN and VHL groups, keeping the E3 ligand class fixed.
We also assess whether nearby molecules in chemical space tend to share the same POI label. Observed nearest-neighbor purity is compared with a null distribution obtained by shuffling POI labels within each E3 group. This preserves POI frequencies while breaking the relationship between POI identity and molecular structure.

The analysis shows that molecules targeting the same POI are more similar on average, even within fixed E3 groups. Their local chemical neighborhoods also show strong POI enrichment.  At the same time, individual POIs can span multiple chemotypes rather than forming a single homogeneous cluster. PROTAC chemical space is therefore globally heterogeneous but locally organized by POI identity. This motivates evaluating conditional generation using both likelihood-based sensitivity to the protein condition and local enrichment in held-out target-specific chemical neighborhoods, rather than similarity to a single chemical centroid.

\begin{figure*}[!t]
\centering
\includegraphics[width=\textwidth]{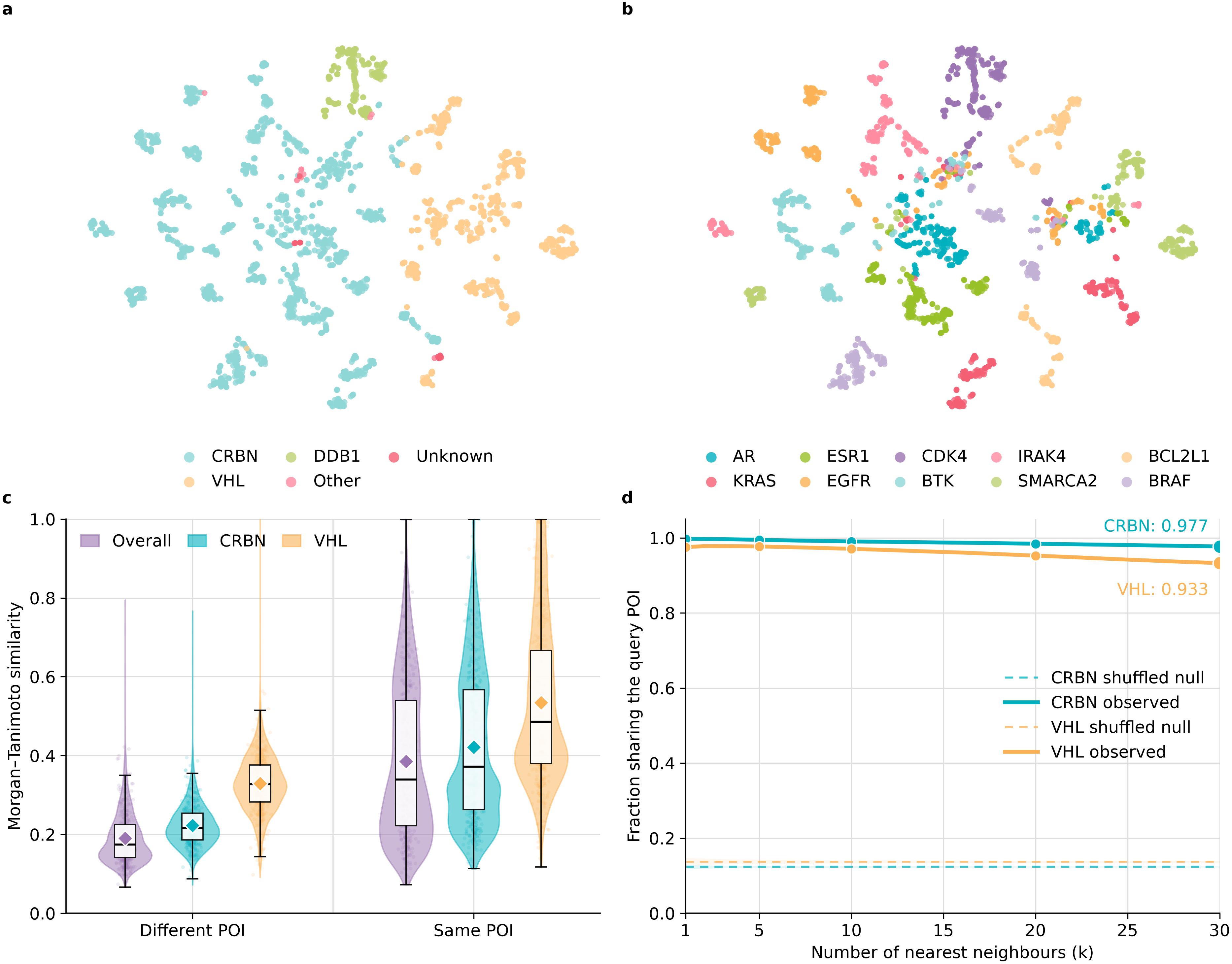}
\caption{
\textbf{POI-specific PROTAC chemistry is globally heterogeneous but locally organized.}
All analyses use 2048-bit Morgan fingerprints with radius 2.
\textbf{(a,b)} The same two-dimensional embedding of PROTAC chemical space, colored by \textbf{(a)} E3-ligase identity and \textbf{(b)} POI identity. Both panels contain 3,000 molecules, with 300 sampled from each of the 10 most represented POIs. E3 identity broadly structures chemical space, while individual POIs show local enrichment across distinct regions.
\textbf{(c)} Morgan--Tanimoto similarity between same-POI and different-POI molecular pairs across the seen POIs. We sample 10,000 pairs of each type. Same-POI similarity is higher overall (0.385 versus 0.190), within CRBN (0.421 versus 0.223), and within VHL (0.535 versus 0.329), indicating that the POI-specific similarity signal persists after controlling for E3 identity.
\textbf{(d)} Local POI purity within CRBN and VHL chemical space. Purity measures the fraction of the $k$ nearest neighbors sharing the query's POI, with $k=1$--30. Solid lines show observed purity, while dashed lines represent the null obtained from 300 within-E3 label permutations. At $k=30$, observed purity remains high for CRBN (0.977) and VHL (0.933), well above the shuffled-label null. Strong local enrichment despite moderate global similarity indicates that individual POIs can span multiple locally coherent chemotypes.
}
\label{fig:poi_signal}
\end{figure*}

The full analysis is detailed in Appendix~\ref{app:poi_structure}, with results shown in Figure~\ref{fig:poi_signal}.

\paragraph{Generation quality.}
We evaluate generation quality using validity, uniqueness, and novelty, defined as
\begin{equation}
\mathrm{V}
=
\frac{|\mathcal{V}|}{|\mathcal{G}|},
\qquad
\mathrm{U}
=
\frac{|\mathcal{U}|}{|\mathcal{V}|},
\qquad
\mathrm{Nov}
=
\frac{
\left|
\left\{
g\in\mathcal{U}
:
g\notin\mathcal{R}_{\mathrm{train}}
\right\}
\right|
}{
|\mathcal{U}|
}.
\end{equation}

Here, $\mathcal{R}_{\mathrm{train}}$ denotes the set of unique molecules in the training data. Validity measures the fraction of valid generated molecules, uniqueness the fraction of valid molecules that are distinct, and novelty the fraction of unique valid molecules that are not in the training set.

For conditional generation, these metrics are computed for each POI $p$ using $(\mathcal{G}_p,\mathcal{V}_p,\mathcal{U}_p)$. Novelty is measured against all molecules in the training set, regardless of POI identity. A generated molecule is therefore novel only if it does not appear anywhere in the training data, providing a stricter and more conservative measure of novelty.

We measure structural diversity among unique valid generated molecules using internal diversity defined as
\begin{equation}
\mathrm{IntDiv}
=
1-
\binom{|\mathcal{U}|}{2}^{-1}
\sum_{i<j}
T(f_i,f_j),
\end{equation}
Here, $f_i$ is the Morgan fingerprint of molecule $i$ and $T(f_i,f_j)$ is the Tanimoto similarity between fingerprints. Higher internal diversity indicates greater structural variation among generated molecules.

We further characterize generated molecules using molecular weight, topological polar surface area (TPSA), LogP, heavy-atom count, rotatable-bond count, aromatic and non-aromatic ring counts, and hydrogen-bond acceptor and donor counts.

Fingerprint and molecular-property calculations are detailed in Appendix~\ref{app:evaluation_details}.

\paragraph{POI-warhead analysis.}
Whole-molecule diversity does not necessarily reflect diversity at the POI-binding end of a PROTAC. We therefore use PROTAC-Splitter~\citep{Ribes2026PROTACSplitter} to decompose unique valid generated PROTACs into warhead, linker, and E3 ligand components. We then evaluate the predicted warheads per POI. 

The extracted warheads are converted to canonical parent structures. Let $\mathcal{W}_p$ denote the set of successfully processed parent warheads from  PROTACs generated for POI $p$, and $\mathcal{W}_p^*$ be its set of unique canonical warheads.
Let $\mathcal{R}^{W}_p$ denote the set of unique canonical parent warheads extracted from training PROTACs targeting the same POI $p$.

Warhead uniqueness and novelty are defined as
\begin{equation}
\mathrm{WUniq}(p)
=
\frac{|\mathcal{W}^{*}_p|}{|\mathcal{W}_p|},
\qquad
\mathrm{WNov}(p)
=
\frac{
|\mathcal{W}^{*}_p\setminus\mathcal{R}^{W}_p|
}{
|\mathcal{W}^{*}_p|
},
\end{equation}
with internal diversity over unique generated warheads,
\begin{equation}
\mathrm{WDiv}(p)
=
1-
\binom{|\mathcal{W}^{*}_p|}{2}^{-1}
\sum_{i<j}
T(w_i,w_j).
\end{equation}
where $T(w_i,w_j)$ denotes the Tanimoto similarity between warhead fingerprints.

PROTAC-Splitter processing, parent-warhead construction, and quality control are detailed in Appendix~\ref{app:warhead_details}.

\paragraph{Conditional fidelity.}
Generation quality does not establish whether the decoder responds to the protein condition. We therefore assess conditional fidelity using two complementary measures.

First, we examine whether held-out molecules receive higher likelihood under their correct POI condition than under competing conditions. For a held-out molecule $x$ and POI $p$, the length-normalized negative log-likelihood is defined as
\begin{equation}
\mathrm{NLL}(x\mid p)
=
-\frac{1}{T_x}
\sum_{t=1}^{T_x}
\log
P_{\theta}
\left(
s_t
\mid
s_{<t},
\mathbf{z}_p
\right).
\end{equation}
Let $\mathcal{T}$ denote the set of POIs in the current evaluation. For a held-out molecule $x$, let $\mathcal{P}(x)$ denote the subset of POIs that are known targets of $x$. Conditional preference is measured by
\begin{equation}
\Delta_{\mathrm{NLL}}(x,p)
=
\frac{1}
{|\mathcal{T}\setminus\mathcal{P}(x)|}
\sum_{q\in\mathcal{T}\setminus\mathcal{P}(x)}
\mathrm{NLL}(x\mid q)
-
\mathrm{NLL}(x\mid p).
\end{equation}
A positive value indicates higher likelihood under the correct POI condition than under competing conditions, on average. If a molecule has multiple valid POIs, all of them are excluded from the competing set to avoid treating correct conditions as incorrect.

Second, we measure POI enrichment among the nearest held-out reference neighbors of generated molecules. Queries are restricted to valid, unique, training-novel molecules. For molecule $g$ generated under POI $p$, let $\mathcal{N}_k(g)$ denote its $k$ nearest molecules in the held-out test reference pool by Morgan--Tanimoto similarity. POI purity is
\begin{equation}
\rho_k(g,p)
=
\frac{1}{k}
\sum_{r\in\mathcal{N}_k(g)}
\mathbf{1}
\left[
p\in\mathcal{P}(r)
\right],
\end{equation}
We average this score over queries generated for POI $p$ and report $k\in\{5,10,20\}$.

To provide context for generated purity, we also measure the local POI organization of the held-out reference molecules using leave-one-out nearest-neighbor analysis. Let $\mathcal{R}_p$ denote the unique held-out reference molecules targeting POI $p$, and $\mathcal{N}_k^{-x}(x)$ denote the $k$ nearest reference molecules to $x$, excluding $x$ itself. Reference purity is defined as
\begin{equation}
\mathrm{RefPurity}_k(p)
=
\frac{1}{|\mathcal{R}_p|}
\sum_{x\in\mathcal{R}_p}
\frac{1}{k}
\sum_{r\in\mathcal{N}_k^{-x}(x)}
\mathbf{1}\left[p\in\mathcal{P}(r)\right].
\end{equation}
This provides a reference point for interpreting generated neighborhood purity.

Reference-pool construction and multi-label handling are described in Appendix~\ref{app:conditional_fidelity}.

\paragraph{Baseline.}
We use the unconditional model as a target-agnostic baseline to quantify the contribution of protein conditioning to POI-associated chemical enrichment. For each POI $p$, we compare the expected $k$-NN purity under conditional and unconditional generation as
\begin{equation}
\Delta_k(p)
=
\underbrace{
\mathbb{E}_{g\sim P_{\theta}(\cdot\mid\mathbf{z}_p)}
[\rho_k(g,p)]
}_{\text{Conditional}}
-
\underbrace{
\mathbb{E}_{g\sim P_{\phi}}
[\rho_k(g,p)]
}_{\text{Unconditional}}.
\end{equation}
Both expectations are estimated using valid, unique, training-novel generated molecules against the same held-out reference. A positive $\Delta_k(p)$ indicates greater local enrichment for POI $p$ under conditional generation. We report $\Delta_k(p)$ for $k\in\{5,10,20\}$.

\section{Results}
\label{sec:results}

\subsection{Molecular pretraining and unconditional generation}
\label{sec:unconditional_results}
Pretraining on PROTAC-like molecules produces a molecular prior capable of generating valid, novel, and diverse structures. From 10,000 generated samples, the pretrained model achieves $98.18\%$ validity, $91.99\%$ uniqueness, $66.67\%$ novelty, and an internal diversity of $0.836$.

Fine-tuning this molecular prior on PROTAC data without protein conditioning produces the unconditional model. Across three seeds, with 200,000 generated molecules per seed, the unconditional model achieves $(99.05 \pm 0.01)\%$ validity, $(47.13 \pm 0.14)\%$ uniqueness, $(89.54 \pm 0.08)\%$ novelty, and an internal diversity of $0.792 \pm 0.001$.

\subsection{Target-conditioned generation for seen POIs}
\label{sec:seen_poi_results}

Table~\ref{tab:seen_poi_results} summarizes generation quality and conditional fidelity for the seen POIs. We generate 10,000 molecules per target and report metrics averaged across three sampling seeds.

\begin{table}[h!]
\caption{
\textbf{Target-wise generation quality and conditional fidelity across the seen POIs.}
Generation metrics and nearest-neighbor purity are reported as mean (standard deviation) across three generation seeds. Best per-target values in each column are shown in bold.
}
\label{tab:seen_poi_results}
\centering
\scriptsize
\setlength{\tabcolsep}{6pt}
\resizebox{\linewidth}{!}{%
\begin{tabular}{lcccccccc}
\hline
\textbf{POI} &
\textbf{Validity (\%)} &
\textbf{Uniqueness (\%)} &
\textbf{Novelty (\%)} &
\textbf{Diversity} &
\textbf{$\Delta_{\mathrm{NLL}}$} &
\textbf{5-NN (\%)} &
\textbf{10-NN (\%)} &
\textbf{20-NN (\%)} \\
\hline
AR      & 99.65 (0.07) & 48.64 (0.53) & 79.38 (0.09) & 0.689 (0.004) & 0.278 & 97.89 (0.15) & 97.32 (0.16) & 96.33 (0.15) \\
ESR1    & 99.53 (0.11) & 41.12 (0.38) & 80.54 (0.13) & 0.656 (0.001) & 0.366 & 97.94 (0.20) & 97.12 (0.25) & 94.78 (0.23) \\
CDK4    & 99.58 (0.05) & 30.60 (0.14) & 81.86 (0.20) & 0.566 (0.006) & \textbf{0.618} & 95.17 (0.10) & 93.62 (0.23) & 92.66 (0.36) \\
KRAS    & 99.46 (0.06) & 49.80 (0.26) & 85.59 (0.18) & 0.641 (0.004) & 0.498 & 97.91 (0.08) & 97.18 (0.06) & 95.56 (0.09) \\
EZH2    & 98.40 (0.01) & 31.50 (0.55) & 94.57 (0.14) & 0.629 (0.003) & 0.358 & 82.38 (0.25) & 74.63 (0.13) & 56.15 (0.30) \\
BTK     & 97.85 (0.15) & 43.06 (0.19) & 87.63 (0.31) & 0.716 (0.003) & 0.428 & 88.58 (0.21) & 85.50 (0.20) & 81.07 (0.21) \\
BRD4    & 97.16 (0.07) & 41.63 (0.30) & 91.02 (0.26) & 0.720 (0.006) & 0.287 & 66.19 (0.77) & 59.97 (0.82) & 50.83 (0.73) \\
BRAF    & 99.70 (0.01) & 28.77 (0.24) & 88.36 (0.14) & 0.612 (0.001) & 0.355 & 92.59 (0.27) & 90.22 (0.33) & 80.50 (0.32) \\
SMARCA4 & 98.62 (0.02) & 41.18 (0.38) & 90.29 (0.27) & 0.671 (0.001) & 0.347 & 27.37 (1.20) & 26.29 (1.17) & 23.59 (0.98) \\
MAP4K1  & 98.29 (0.11) & 33.90 (0.19) & 94.63 (0.11) & \textbf{0.726 (0.003)} & 0.418 & 61.59 (0.74) & 40.43 (0.49) & 22.65 (0.15) \\
IRAK4   & 97.68 (0.11) & \textbf{57.13 (0.68)} & 94.56 (0.07) & 0.701 (0.005) & 0.507 & 93.97 (0.11) & 92.49 (0.18) & 85.74 (0.09) \\
EGFR    & 97.91 (0.10) & 52.05 (0.44) & 89.51 (0.30) & 0.723 (0.005) & 0.381 & 90.86 (0.16) & 85.26 (0.23) & 76.40 (0.34) \\
BCL2L1  & 98.94 (0.05) & 37.25 (0.26) & 88.20 (0.07) & 0.682 (0.003) & 0.422 & 91.71 (0.38) & 88.58 (0.54) & 82.54 (0.67) \\
SMARCA2 & 98.58 (0.06) & 42.12 (0.45) & 89.86 (0.36) & 0.672 (0.003) & 0.327 & 79.30 (0.32) & 73.34 (0.22) & 67.61 (0.11) \\
METTL14 & \textbf{99.84 (0.02)} & 13.74 (0.29) & 81.58 (0.20) & 0.386 (0.001) & 0.564 & \textbf{99.57 (0.14)} & \textbf{99.55 (0.15)} & \textbf{99.41 (0.13)} \\
PTPN1   & 97.55 (0.09) & 24.61 (0.24) & 92.79 (0.13) & 0.589 (0.007) & 0.538 & 81.92 (0.63) & 78.96 (0.65) & 77.22 (0.66) \\
BCL2    & 97.29 (0.16) & 28.87 (0.41) & 93.85 (0.17) & 0.599 (0.010) & 0.394 & 75.60 (0.17) & 71.92 (0.25) & 64.41 (0.48) \\
SOS1    & 99.25 (0.11) & 33.99 (0.64) & \textbf{95.34 (0.07)} & 0.645 (0.001) & 0.356 & 86.51 (0.49) & 74.80 (0.27) & 51.12 (0.14) \\
BCL6    & 99.30 (0.13) & 28.50 (0.64) & 91.04 (0.14) & 0.467 (0.007) & 0.516 & 98.60 (0.09) & 98.52 (0.13) & 98.04 (0.14) \\
MAPT    & 99.61 (0.03) & 30.16 (0.16) & 93.13 (0.03) & 0.562 (0.003) & 0.494 & 89.00 (0.54) & 86.40 (0.54) & 83.60 (0.51) \\
\hline
\textbf{Macro} &
\textbf{98.71 (0.02)} &
\textbf{36.93 (0.04)} &
\textbf{89.19 (0.06)} &
\textbf{0.633 (0.001)} &
\textbf{0.423} &
\textbf{84.73 (0.11)} &
\textbf{80.60 (0.11)} &
\textbf{74.01 (0.14)} \\
\hline
\end{tabular}%
}
\end{table}

Validity remains high across all POIs, with a macro average of $98.71\%$ and a minimum of $97.16\%$. Mean novelty reaches $89.19\%$. Uniqueness and internal diversity vary more substantially, with macro averages of $36.93\%$ and $0.633$, respectively.

Both conditional-fidelity measures show sensitivity to POI identity.
$\Delta_{\mathrm{NLL}}$ is positive for all seen POIs, with a macro average of $+0.423$. Held-out molecules therefore receive lower negative log-likelihood, on average, under their correct POI condition than under competing conditions.

The generated molecules also show strong POI-specific neighborhood enrichment. Macro purity reaches $84.73\%$ at $k=5$, $80.60\%$ at $k=10$, and $74.01\%$ at $k=20$. For comparison, the corresponding leave-one-out reference purity values are $92.27\%$, $86.72\%$, and $78.33\%$ (Appendix Table~\ref{tab:seen_real_gen_knn}). Thus, generated molecules reproduce much of the local POI enrichment observed in the held-out reference. The strength of this enrichment nevertheless varies across targets. At $k=5$, generated purity ranges from $27.37\%$ for SMARCA4 to $99.57\%$ for METTL14.

Warhead analysis further shows that the model generates novel warheads, with a macro average of $86.62\%$ (Appendix Table~\ref{tab:warhead_results}). PROTAC-Splitter yields usable three-component decompositions for $99.99\%$ of unique valid generated PROTACs, averaged across POIs. Of these decompositions, $99.97\%$ yield valid parent warheads (Appendix Table~\ref{tab:splitter_coverage}).

\begin{figure*}[h]
\centering
\includegraphics[width=\textwidth]{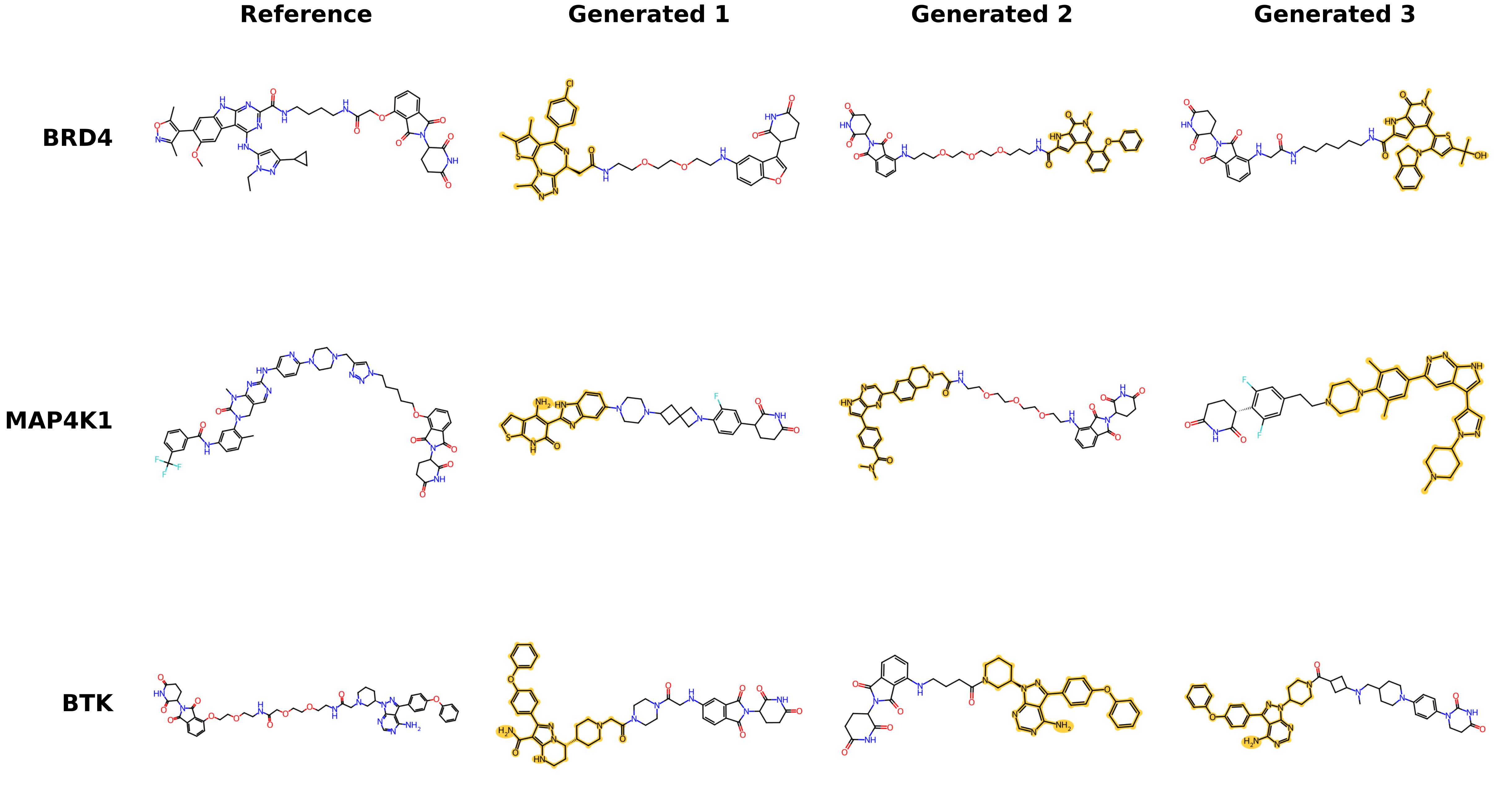}
\caption{
\textbf{Generated PROTACs span distinct warhead scaffolds under the same POI condition.}
Rows show BRD4, MAP4K1, and BTK, each with one reference PROTAC and three generated examples. These targets have the highest warhead internal diversity ($0.869$, $0.835$, and $0.817$, respectively).
Gold highlighting identifies the warhead predicted by PROTAC-Splitter in each generated molecule. Bemis--Murcko scaffolds are extracted from the parent warheads to group molecules by their core structures. For each POI, the generated examples represent three distinct warhead-scaffold groups.
}
\label{fig:warhead_examples}
\end{figure*}

Figure~\ref{fig:warhead_examples} illustrates generated PROTACs for BRD4, MAP4K1, and BTK,  which have the highest warhead internal diversity. The examples show distinct warhead scaffolds generated under the same protein condition.

\subsection{Comparison with the unconditional baseline}
\label{sec:baseline_results}

To quantify the contribution of protein conditioning, we compare TACFormer with the matched unconditional baseline. Both models are evaluated against the same held-out reference pool, and $\Delta_k(p)$ measures their difference in $k$-NN purity for POI $p$ (Section~\ref{sec:evaluation}).

Table~\ref{tab:baseline_enrichment} shows positive improvements for every seen POI at $k=5$, $k=10$, and $k=20$. The macro improvements are $79.71$, $75.58$, and $68.99$ percentage points, respectively. 

At $k=5$, the largest improvements occur for METTL14 and BCL6, reaching $97.05$ and $96.85$ percentage points, respectively. SMARCA4 shows the smallest improvement at $26.45$ percentage points. These positive differences demonstrate that protein conditioning increases enrichment for the requested POI in held-out chemical neighborhoods compared with unconditional generation.

The target-wise reference, conditional, and unconditional purity values are reported in Appendix Table~\ref{tab:seen_real_gen_knn}.

\begin{table}[h!]
\caption{
\textbf{Improvement in POI-neighborhood purity over unconditional generation.}
$\Delta_k(p)$ measures the difference between conditional and unconditional $k$-NN purity for each POI. Values are reported in percentage points as mean (standard deviation) across three generation seeds.
}
\label{tab:baseline_enrichment}
\centering
\scriptsize
\setlength{\tabcolsep}{6pt}
\begin{tabular}{lccc}
\hline
\textbf{POI} &
\textbf{$\Delta_5$ (pp)} &
\textbf{$\Delta_{10}$ (pp)} &
\textbf{$\Delta_{20}$ (pp)} \\
\hline
AR       & 75.14 (0.02) & 74.14 (0.10) & 72.32 (0.10) \\
ESR1     & 88.03 (0.18) & 86.86 (0.24) & 84.10 (0.21) \\
CDK4     & 89.38 (0.06) & 87.94 (0.15) & 87.01 (0.27) \\
KRAS     & 87.95 (0.06) & 87.13 (0.05) & 85.41 (0.10) \\
EZH2     & 80.82 (0.27) & 73.01 (0.12) & 54.64 (0.28) \\
BTK      & 82.32 (0.23) & 79.23 (0.20) & 74.75 (0.20) \\
BRD4     & 60.87 (0.74) & 54.94 (0.80) & 46.24 (0.73) \\
BRAF     & 89.65 (0.33) & 87.16 (0.39) & 77.41 (0.34) \\
SMARCA4  & 26.45 (1.19) & 25.38 (1.16) & 22.70 (0.97) \\
MAP4K1   & 60.05 (0.74) & 39.16 (0.50) & 21.58 (0.15) \\
IRAK4    & 90.26 (0.11) & 88.69 (0.17) & 82.00 (0.05) \\
EGFR     & 76.64 (0.22) & 71.84 (0.27) & 63.90 (0.37) \\
BCL2L1   & 89.20 (0.38) & 86.02 (0.56) & 79.96 (0.67) \\
SMARCA2  & 76.44 (0.31) & 70.59 (0.21) & 64.92 (0.06) \\
METTL14  & \textbf{97.05 (0.12)} & \textbf{96.66 (0.14)} & \textbf{96.13 (0.12)} \\
PTPN1    & 80.92 (0.61) & 77.96 (0.63) & 76.20 (0.64) \\
BCL2     & 74.18 (0.19) & 70.46 (0.26) & 63.04 (0.48) \\
SOS1     & 85.23 (0.48) & 73.44 (0.26) & 49.83 (0.13) \\
BCL6     & 96.85 (0.10) & 96.66 (0.13) & 96.01 (0.15) \\
MAPT     & 86.76 (0.56) & 84.34 (0.55) & 81.62 (0.51) \\
\hline
\textbf{Macro}
& \textbf{79.71 (0.11)}
& \textbf{75.58 (0.11)}
& \textbf{68.99 (0.14)} \\
\hline
\end{tabular}
\end{table}

\subsection{Zero-shot generation}
\label{sec:zero_shot_results}

Table~\ref{tab:zero_shot_results} reports generation quality and conditional fidelity for the four zero-shot POIs. Generation quality remains high overall, with macro validity of $91.76\%$, novelty of $98.19\%$, and internal diversity of $0.712$. Validity is notably lower for WDR5 ($75.44\%$) than for the other targets, which reach at least $95.33\%$.

Likelihood-based conditional fidelity differs across the four proteins. WDR5 and DHFR both show negative $\Delta_{\mathrm{NLL}}$ values of $-0.089$, indicating that their correct POI conditions are not preferred over competing conditions on average. In contrast, FGFR3 and MAP3K13 show positive values of $+0.080$ and $+0.096$, respectively. These results suggest a relationship between protein-representation similarity and likelihood preference among the selected zero-shot targets.

Nearest-neighbor fidelity reveals a different pattern.Macro purity reaches $30.52\%$ at $k=5$, $31.55\%$ at $k=10$, and $32.29\%$ at $k=20$. At $k=5$, WDR5 achieves the highest purity ($56.23\%$), followed by DHFR ($31.05\%$), FGFR3 ($20.26\%$), and MAP3K13 ($14.57\%$). By comparison, the held-out reference molecules exhibit strong local POI enrichment, with a macro 5-NN purity of $99.25\%$ (Appendix Table~\ref{tab:unseen_real_gen_knn}).

\begin{table*}[h!]
\caption{
\textbf{Generation quality and conditional fidelity for the four zero-shot POIs.}
WDR5, DHFR, FGFR3, and MAP3K13 are excluded entirely from conditional training and provided to the model only at inference. Generation metrics and nearest-neighbor purity are reported as mean (standard deviation) across three generation seeds.
}
\label{tab:zero_shot_results}
\centering
\scriptsize
\setlength{\tabcolsep}{5pt}
\begin{tabular}{lcccccccc}
\hline
\textbf{POI} &
\textbf{Validity (\%)} &
\textbf{Uniq. (\%)} &
\textbf{Novelty (\%)} &
\textbf{Diversity} &
\textbf{$\Delta_{\mathrm{NLL}}$} &
\textbf{5-NN (\%)} &
\textbf{10-NN (\%)} &
\textbf{20-NN (\%)} \\
\hline
WDR5
& 75.44 (0.53) & \textbf{99.21 (0.08)} & \textbf{99.96 (0.01)}
& \textbf{0.737 (0.002)} & $-0.089$
& \textbf{56.23 (0.10)} & \textbf{55.35 (0.35)} & \textbf{50.58 (0.44)} \\

DHFR
& 95.33 (0.03) & 88.59 (0.20) & 99.57 (0.01)
& 0.669 (0.003) & $-0.089$
& 31.05 (0.32) & 37.06 (0.28) & 45.18 (0.14) \\

FGFR3
& 96.96 (0.04) & 45.95 (0.57) & 97.31 (0.08)
& 0.716 (0.003) & 0.080
& 20.26 (0.15) & 20.51 (0.07) & 20.70 (0.05) \\

MAP3K13
& \textbf{99.29 (0.12)} & 36.23 (0.22) & 95.90 (0.25)
& 0.727 (0.004) & \textbf{0.096}
& 14.57 (0.46) & 13.27 (0.31) & 12.71 (0.12) \\
\hline
\textbf{Macro}
& \textbf{91.76 (0.11)}
& \textbf{67.50 (0.15)}
& \textbf{98.19 (0.04)}
& \textbf{0.712 (0.001)}
& \textbf{$-0.0005$}
& \textbf{30.52 (0.06)}
& \textbf{31.55 (0.11)}
& \textbf{32.29 (0.14)} \\
\hline
\end{tabular}
\end{table*}

Thus, the positive likelihood preference observed for FGFR3 and MAP3K13 does not translate into strong enrichment in their held-out chemical neighborhoods. These results suggest that similarity in ESM-2 embedding space may support likelihood-based generalization without ensuring comparable improvement in generated chemical distributions. Zero-shot generalization remains limited relative to the seen-POI results.

\subsection{Invalid SMILES failure modes}
\label{sec:invalid_smiles_results}

The model generates 2,582 invalid SMILES across the seen POI conditions. The most frequent recorded failures are kekulization errors ($43.45\%$), unclosed ring labels ($42.41\%$), and incorrect parentheses ($10.34\%$). Valence or atomic-property violations account for $2.13\%$, and duplicate ring-closure bonds account for $1.16\%$.

Failure patterns also differ across POIs. Some conditions produce more aromaticity-related errors, while others show more ring-closure or branching-syntax failures. These findings indicate that most invalid SMILES arise from difficulties in maintaining valid ring connectivity and aromatic structures during autoregressive generation.

\section{Discussion}
\label{sec:discussion}
The results demonstrate that protein conditioning enables TACFormer to capture POI-specific patterns in PROTAC chemical space beyond those learned without target information. For the seen POIs, likelihood preference and generated neighborhood purity provide complementary evidence that the protein condition influences both how the model assigns likelihood to molecules and the chemical distributions it generates.

The unconditional baseline provides an important comparison. It generates highly valid, novel, and diverse PROTACs but achieves lower neighborhood purity. This shows that learning the overall PROTAC distribution is not sufficient to reproduce the chemical patterns that distinguish individual targets. Because both models are fine-tuned on matched molecular data, the improvement achieved by TACFormer supports the contribution of protein conditioning rather than molecular fine-tuning alone.

\paragraph{Conditional fidelity depends on the structure of the reference chemical space.}
Generated $k$-NN purity varies considerably across POIs, but its value must be interpreted in the context of the corresponding reference chemistry. Held-out reference molecules also differ in neighborhood purity. For some targets, local neighborhoods remain strongly enriched for the same POI, whereas for others, they become increasingly mixed as $k$ increases. Consequently, generated purity alone does not fully characterize how well the model captures the chemical organization of the reference data.

MAP4K1 illustrates this distinction. Its generated 5-NN purity reaches $61.59\%$, below targets such as AR or METTL14, but its held-out reference purity is only $70.00\%$. At $k=20$, generated and reference purity are nearly identical ($22.65\%$ and $23.33\%$, respectively). The lower purity therefore largely reflects the organization of the available MAP4K1 reference chemistry rather than a failure to reproduce it. SMARCA4 shows a different pattern. Its reference 5-NN purity reaches $77.39\%$, whereas generated purity is only $27.37\%$. Here, the large reference--generation gap indicates that TACFormer reproduces the local POI enrichment of the reference molecules less consistently. Nevertheless, its generated purity still exceeds the unconditional baseline by $26.45$ percentage points, showing that protein conditioning still improves enrichment relative to target-agnostic generation.

Comparing generated and reference purity therefore provides important context for evaluating conditional fidelity. For several targets, including AR, ESR1, EGFR, BCL2L1, METTL14, BCL6, and MAPT, generated purity closely approaches the corresponding reference values. For KRAS and SOS1, it even exceeds reference purity at larger neighborhood sizes. However, this does not imply greater biological target specificity or more complete coverage of relevant chemistry. Instead, the model may preferentially generate molecules from highly enriched regions while sampling other chemotypes less frequently. This would produce high local purity without necessarily covering the broader reference chemical space. Neighborhood purity therefore measures local POI enrichment, not the completeness of chemotype coverage.

This distinction is important when interpreting the diversity of generated molecules. High neighborhood purity can coexist with limited chemical diversity if generation concentrates within a few POI-enriched regions. METTL14 illustrates this possibility. It achieves $99.57\%$ generated 5-NN purity but only $13.74\%$ uniqueness and $0.386$ internal diversity. BCL6 similarly combines high neighborhood purity with low diversity. These findings are consistent with generation concentrated within a narrower range of chemical structures, although they do not establish which reference chemotypes are covered. Conversely, a more diverse generated distribution may span several chemical regions with different levels of local POI enrichment.

The relationship between internal diversity and conditional fidelity also depends on the evaluation measure. Across the seen POIs, internal diversity is negatively correlated with $\Delta_{\mathrm{NLL}}$ ($R^2=0.375$, $p=0.0041$), whereas its relationship with 5-NN purity is weaker and not statistically significant ($R^2=0.118$, $p=0.139$). Importantly, $\Delta_{\mathrm{NLL}}$ measures likelihood preference for held-out molecules, while internal diversity measures variation among generated molecules. Their correlation therefore does not establish that reducing molecular diversity improves conditional fidelity. Instead, these findings show that likelihood preference, local POI enrichment, and structural diversity capture different properties of generation and should be interpreted together.

Target-wise variation in conditional fidelity may also reflect differences in the amount and composition of training data. The number of training examples varies across the seen POIs (Appendix~\ref{app:data_splits}), and individual targets can span multiple chemotypes. TACFormer must therefore learn from unequal amounts of data and chemical distributions with different levels of heterogeneity. Training-set size alone does not appear to explain the observed variation, as some targets with few examples achieve strong conditional fidelity while others while others with more examples remain difficult. These observations suggest that both data availability and chemical heterogeneity may jointly contribute to target-wise performance.

\paragraph{Protein conditioning supports diverse warhead generation.}

The warhead analysis shows that structural novelty extends to the warhead component. Macro POI-warhead novelty reaches $86.62\%$, and several targets show substantial warhead diversity. This indicates that TACFormer can also explore new structures at the POI-binding end of the molecule rather than relying exclusively on warheads observed for the same target during training.
 
However, warhead novelty alone does not establish whether generated PROTACs reflect the chemistry of the intended target. SMARCA4, for example, achieves $95.87\%$ warhead novelty but only $27.37\%$ generated 5-NN purity. This contrast shows that generating novel warheads does not necessarily produce molecules enriched in the chemical neighborhoods of the requested target. Warhead novelty should therefore be evaluated alongside conditional fidelity to assess both structural exploration and how effectively protein conditioning guides molecular generation.

\paragraph{Protein conditioning generalizes partially to unseen POIs.}
The zero-shot results reveal a distinction between likelihood-based preference and the ability to generate chemistry characteristic of an unseen target. FGFR3 and MAP3K13, whose ESM-2 representations are similar to those of training proteins, show positive likelihood preference. However, this preference does not translate into strong enrichment in their held-out chemical neighborhoods. Conversely, WDR5 shows negative likelihood preference but achieves the highest generated neighborhood purity among the four zero-shot POIs. These findings indicate that likelihood-based transfer and generated neighborhood enrichment do not necessarily follow the same pattern.

One possible explanation is that similarity in ESM-2 embedding space supports partial transfer of learned chemical preferences but does not sufficiently constrain generation toward the chemotypes of a new POI. This suggests that likelihood-based transfer may be easier to achieve than reproducing the corresponding chemical distributions. The high reference neighborhood purity of the zero-shot POIs indicates strong local chemical organization in the available held-out data, but TACFormer captures this structure only partially. 

\subsection{Limitations and future directions.}

The current evaluation demonstrates that protein conditioning guides generation toward chemistry characteristic of the requested POI, but it does not establish biological activity. Neither $\Delta_{\mathrm{NLL}}$ nor neighborhood purity provides evidence of target binding, degradation potency, or selectivity. A generated molecule outside the observed reference neighborhoods may still be active, and similarity to known PROTACs does not guarantee degradation activity. Future work could explore degradation-activity prediction using PROTACs with measured outcomes, such as $\mathrm{DC}_{50}$ and $D_{\max}$.  Such a predictor could estimate degradation potency and efficacy for generated molecules, complementing the current work by allowing generated PROTACs to be ranked according to their predicted degradation potency and efficacy.

Another limitation concerns how TACFormer represents the protein condition. Each POI is encoded using a single mean-pooled ESM-2 embedding. Although this provides a compact representation of the protein sequence, pooling compresses residue-level information into one vector and may obscure sequence regions important for ligand recognition. This may be particularly limiting to zero-shot generation, where no molecular examples for the unseen targets are available during training to compensate for information lost during pooling.

Future work could investigate residue-level ESM-2 conditioning using attention mechanisms that allow the molecular decoder to access information from individual protein positions during SMILES generation. This would require modifications to the current architecture and evaluation of its effect on conditional generation. Protein structural information could also be explored as a complementary approach for evaluating and prioritizing generated PROTACs, where suitable data are available. These directions could help assess whether more detailed protein information improves generation and generalization to unseen POIs.

\section{Conclusion}
\label{sec:conclusion}
TACFormer demonstrates that protein sequence conditioning can steer PROTAC generation toward chemical patterns characteristic of the requested target. Across the seen POIs, positive likelihood preference, strong enrichment in held-out chemical neighborhoods, and consistent improvements over the matched unconditional baseline provide complementary evidence that protein information guides molecular generation beyond the general PROTAC prior. The results also highlight that conditional fidelity, chemical diversity, chemotype coverage, and warhead novelty represent distinct aspects of generation that should be evaluated together rather than in isolation.

The zero-shot experiments reveal an important limitation. Positive likelihood preference for some unseen POIs does not consistently translate into strong enrichment in their reference chemical neighborhoods. Similarity between protein representations may support likelihood-based transfer but does not guarantee that generated molecules reproduce the chemistry observed for unseen targets.

TACFormer provides a foundation for sequence-conditioned PROTAC generation, while improving zero-shot generalization and integrating degradation-activity prediction remain important directions for future research.

\section*{Author contributions}

\noindent\textbf{YG:} Conceptualization, data curation, methodology, software, analysis, investigation, validation, visualization, writing -- original draft, and writing -- review \& editing. \textbf{RM:} Conceptualization, methodology, supervision, funding acquisition, resources, writing -- original draft, and writing -- review \& editing.

% ================================================================
% ACKNOWLEDGMENTS
% ================================================================

\begin{ack}
YG and RM acknowledge funding from the Wallenberg AI, Autonomous Systems and Software Program (WASP), supported by the Knut and Alice Wallenberg Foundation. Computational resources were provided by the National Academic Infrastructure for Supercomputing in Sweden (NAISS), funded by the Swedish Research Council.
\end{ack}

% ================================================================
% SOFTWARE AND DATA AVAILABILITY
% ================================================================

\section*{Software and data availability}

Code is available at
\url{https://github.com/yossragh/TACFormer}.

% ================================================================
% REFERENCES
% ================================================================

\bibliographystyle{unsrtnat}
\bibliography{latex/references}

% ================================================================
% APPENDIX
% ================================================================
\newpage
\appendix

\section{Model and implementation details}
\label{app:model_details}

\subsection{SMILES tokenization and augmentation}
\label{app:tokenization}
Molecules are represented as SMILES strings and tokenized using the same molecular tokenizer during pretraining, unconditional fine-tuning, and conditional fine-tuning. Each sequence begins with a beginning-of-sequence (BOS) token and ends with an end-of-sequence (EOS) token. For a tokenized sequence
\begin{equation}
[\mathrm{BOS},s_1,\ldots,s_T,\mathrm{EOS}],
\end{equation}
causal language modeling uses the shifted input--target pair
\begin{equation}
\mathbf{x}
=
[\mathrm{BOS},s_1,\ldots,s_T],
\qquad
\mathbf{y}
=
[s_1,\ldots,s_T,\mathrm{EOS}].
\end{equation}

Sequences are limited to 320 tokens. Within each minibatch, input sequences are right-padded. The corresponding padding positions in the targets are assigned the ignore index $-100$ and excluded from the cross-entropy loss.

SMILES randomization is performed on the fly during training.  Randomized SMILES provide alternative traversals of the same molecular graph, preserving molecular identity while exposing the decoder to different sequence representations. Randomization is applied only to the training data; validation examples retain fixed SMILES representations. For conditional training, the randomized SMILES remains paired with the original POI embedding.

\subsection{Protein sequence representation}
\label{app:protein_representation}
For each POI, we retrieve the amino-acid sequence from UniProtKB using its UniProt accession. Protein representations are computed using the pretrained \texttt{esm2\_t33\_650M\_UR50D} model~\citep{lin2023evolutionary}, which consists of 33 transformer layers, 650M parameters, and a hidden dimension of 1280.

For a POI $p$ with amino-acid sequence
\begin{equation}
\mathbf{a}_p=(a_1,\ldots,a_L),
\end{equation}
ESM-2 produces contextual residue representations. We exclude the initial and terminal special-token representations, retaining one vector per amino-acid residue,
\begin{equation}
\mathbf{H}_p
=
[\mathbf{h}_1,\ldots,\mathbf{h}_L]^\top
\in
\mathbb{R}^{L\times1280}.
\end{equation}

We obtain a fixed-length POI embedding by mean pooling over residues,
\begin{equation}
\mathbf{z}_p
=
\frac{1}{L}
\sum_{i=1}^{L}
\mathbf{h}_i
\in
\mathbb{R}^{1280}.
\end{equation}

Mean pooling removes the variable sequence-length dimension while preserving the 1280-dimensional feature space. The resulting vector $\mathbf{z}_p$ is supplied to the conditional molecular model.

ESM-2 is used as a frozen feature extractor. Protein embeddings are precomputed in FP32, stored by UniProt accession, and reused during molecular-model training and generation. ESM-2 parameters are not updated during conditional fine-tuning.

\subsection{Long-protein encoding}
\label{app:long_sequences}

The ESM-2 model is applied to windows containing at most 1022 amino-acid residues. Proteins satisfying
\begin{equation}
L\leq1022
\end{equation}
are encoded in a single pass. Longer sequences are divided into overlapping windows of maximum length
\begin{equation}
W=1022
\end{equation}
with an overlap of
\begin{equation}
O=256
\end{equation}
residues. The resulting stride is
\begin{equation}
S=W-O=766.
\end{equation}

Overlapping windows provide additional sequence context for residues near window boundaries. Each window is encoded independently by ESM-2, and the resulting residue representations are stitched together before mean pooling.

For adjacent windows, the 256-residue overlap is divided into two equal parts. We retain the representations from the earlier window for the first 128 residues and from the later window for the remaining 128 residues. This favors representations computed farther from window boundaries, where more surrounding sequence context is available. The resulting representation contains exactly one vector per residue.

The reconstructed representation must satisfy
\begin{equation}
\operatorname{rows}(\mathbf{H}_p)=L,
\end{equation}
so that it contains exactly one row for each residue in the original sequence. Mean pooling is then applied to the reconstructed representation:
\begin{equation}
\mathbf{z}_p
=
\frac{1}{L}
\sum_{i=1}^{L}
\mathbf{H}_{p,i}.
\end{equation}

We do not average window-level embeddings, because doing so would weight windows rather than individual residues and could overrepresent shorter terminal windows. Instead, each residue contributes exactly once to the final protein embedding.

\subsection{Decoder architecture}
\label{app:decoder_details}
The molecular model is a decoder-only transformer with eight pre-LayerNorm blocks, a hidden dimension $d_{\mathrm{model}}=512$, eight attention heads, a feed-forward expansion ratio of four, and dropout probability $0.05$. Within each block, $\operatorname{LN}_1$ and $\operatorname{LN}_2$ denote the layer-normalization operations preceding the attention and MLP sublayers, respectively.

For token $s_t$ at sequence position $t$, the input representation is the sum of learned token and positional embeddings,
\begin{equation}
\mathbf{x}_t^{(0)}
=
\mathbf{E}_{\mathrm{tok}}(s_t)
+
\mathbf{E}_{\mathrm{pos}}(t).
\end{equation}

The sequence representations pass through transformer blocks with causal self-attention. In block $\ell$, the input is first normalized,
\begin{equation}
\mathbf{H}^{(\ell)}
=
\operatorname{LN}_1
\left(
\mathbf{X}^{(\ell)}
\right).
\end{equation}
For one attention head, queries, keys, and values are computed as
\begin{equation}
\mathbf{Q}
=
\mathbf{H}^{(\ell)}\mathbf{W}_Q,
\qquad
\mathbf{K}
=
\mathbf{H}^{(\ell)}\mathbf{W}_K,
\qquad
\mathbf{V}
=
\mathbf{H}^{(\ell)}\mathbf{W}_V.
\end{equation}
Causal scaled dot-product attention is defined as
\begin{equation}
\operatorname{Attn}
(\mathbf{Q},\mathbf{K},\mathbf{V})
=
\operatorname{softmax}
\left(
\frac{\mathbf{QK}^{\top}}{\sqrt{d_h}}
+
\mathbf{M}
\right)
\mathbf{V},
\end{equation}
where $d_h=64$ is the head dimension and $\mathbf{M}$ is the causal mask that prevents attention to future tokens.

The eight attention-head outputs are concatenated and projected back to the model dimension. We denote this operation by $\operatorname{MHA}$. The attention residual of block $\ell$ is
\begin{equation}
\mathbf{A}^{(\ell)}
=
\mathbf{X}^{(\ell)}
+
\operatorname{MHA}
\left(
\operatorname{LN}_1
\left(
\mathbf{X}^{(\ell)}
\right)
\right).
\end{equation}

The position-wise multilayer perceptron (MLP) expands the hidden dimension by a factor of four and uses a GELU activation,
\begin{equation}
\operatorname{MLP}(\mathbf{x})
=
\mathbf{W}_2
\operatorname{GELU}
\left(
\mathbf{W}_1\mathbf{x}
+
\mathbf{b}_1
\right)
+
\mathbf{b}_2.
\end{equation}

For the unconditional decoder, the block output is
\begin{equation}
\mathbf{X}^{(\ell+1)}
=
\mathbf{A}^{(\ell)}
+
\operatorname{MLP}
\left(
\operatorname{LN}_2
\left(
\mathbf{A}^{(\ell)}
\right)
\right).
\end{equation}

After the final transformer block, layer normalization and a linear language-model head produce logits over the SMILES vocabulary. The input token embeddings and output projection share their weights.

\subsection{Protein-to-molecule conditioning}

The conditional model retains the molecular decoder and adds a protein adapter that maps the fixed POI embedding to the decoder hidden space. The pooled ESM-2 embedding
\begin{equation}
\mathbf{z}_p\in\mathbb{R}^{1280}
\end{equation}
is first normalized and linearly projected:
\begin{equation}
\mathbf{c}_p
=
\mathbf{W}_c
\operatorname{LN}_{p}
(\mathbf{z}_p)
+
\mathbf{b}_c,
\qquad
\mathbf{W}_c
\in
\mathbb{R}^{512\times1280},
\end{equation}
yielding
\begin{equation}
\mathbf{c}_p\in\mathbb{R}^{512}.
\end{equation}
Here, $\operatorname{LN}_{p}$ is the protein-adapter normalization module, distinct from $\operatorname{LN}_1$ and $\operatorname{LN}_2$ in the decoder blocks.

The projected protein condition $\mathbf{c}_p$ is computed once per forward pass and shared across all decoder blocks. Each block $\ell$ has a trainable scalar gate $\gamma_\ell$. The gated condition is added after the self-attention residual
\begin{equation}
\widetilde{\mathbf{A}}^{(\ell)}
=
\mathbf{A}^{(\ell)}
+
\tanh(\gamma_\ell)\mathbf{c}_p.
\end{equation}

The protein condition is broadcast across all SMILES token positions. The conditioned block output is
\begin{equation}
\mathbf{X}^{(\ell+1)}
=
\widetilde{\mathbf{A}}^{(\ell)}
+
\operatorname{MLP}
\left(
\operatorname{LN}_2
\left(
\widetilde{\mathbf{A}}^{(\ell)}
\right)
\right).
\end{equation}

The same projected protein embedding is used in every block, while the layer-specific gates control its contribution at each depth.

\subsection{Initialization of the conditional model}

The conditional model is initialized from the pretrained molecular prior. Its token embeddings, positional embeddings, attention layers, feed-forward networks, normalization layers, and language-model head are transferred from the prior checkpoint.

The conditional extension adds the protein layer-normalization module $\operatorname{LN}_{p}$, the $1280\rightarrow512$ linear projection, and one scalar gate in each of the eight decoder blocks. The protein LayerNorm is initialized with unit scale and zero bias, the projection is newly initialized, and all gates are initialized to zero,
\begin{equation}
\gamma_\ell^{(0)}=0.
\end{equation}
Therefore,
\begin{equation}
\tanh(\gamma_\ell^{(0)})\mathbf{c}_p
=
\mathbf{0}.
\end{equation}

At initialization, the protein-conditioning pathway contributes zero to every block, so the conditional model retains the pretrained decoder's hidden-state computation. During conditional fine-tuning, the protein adapter, layer-specific gates, and decoder parameters are optimized to incorporate POI information.

The resulting fine-tuning procedure adapts the pretrained unconditional molecular distribution
\begin{equation}
P_{\theta_0}(\mathbf{s})
\end{equation}
to a POI-conditioned distribution
\begin{equation}
P_{\theta}(\mathbf{s}\mid\mathbf{z}_p),
\end{equation}
where $\theta_0$ denotes the pretrained molecular parameters.

\subsection{Causal language-model objective}

During conditional training, each SMILES sequence is paired with the precomputed embedding of its associated POI. For a minibatch of size $B$, the protein embeddings are stacked into
\begin{equation}
\mathbf{Z}
=
[\mathbf{z}_{p_1},\ldots,\mathbf{z}_{p_B}]^\top
\in
\mathbb{R}^{B\times1280}.
\end{equation}
The $n$th protein embedding is aligned with the $n$th SMILES sequence in the minibatch.

Training uses teacher-forced causal language modeling. The loss is the mean token-level negative log-likelihood over non-padding target positions,
\begin{equation}
\mathcal{L}_{\mathrm{LM}}
=
-
\frac{1}{\sum_{n=1}^{B}T_n}
\sum_{n=1}^{B}
\sum_{t=1}^{T_n}
\log
P_{\theta}
\left(
y_t^{(n)}
\mid
x_{\leq t}^{(n)},
\mathbf{z}_{p_n}
\right),
\end{equation}
where $T_n$ is the number of non-padding target tokens in sequence $n$.

At each position, the decoder receives the ground-truth token prefix and the fixed POI embedding to predict the next token. Padding positions are excluded from the loss. ESM-2 is not optimized; training updates the protein adapter, layer-specific gates, and molecular decoder.

\subsection{Optimization details}
\label{app:optimization_details}
Conditional fine-tuning uses a per-device training batch size of 32, a validation batch size of 32, and no gradient accumulation. The learning rate is $1\times10^{-4}$, with weight decay $0.01$ and gradient clipping at a maximum norm of $1.0$. We use a cosine learning-rate schedule with a warm-up ratio of $0.05$ and bfloat16 mixed precision. The same optimization configuration is used for molecular pretraining and unconditional fine-tuning.

\subsection{Autoregressive generation}

At inference, generation begins with a single BOS token. For the conditional generation, the same POI embedding $\mathbf{z}_p$ is supplied at every autoregressive decoding step.

Given decoder logits $\mathbf{o}_t$ at step $t$, the temperature-scaled sampling distribution is
\begin{equation}
P_{\theta}
\left(
s_t
\mid
s_{<t},
\mathbf{z}_p
\right)
=
\operatorname{softmax}
\left(
\frac{\mathbf{o}_t}{\tau}
\right),
\end{equation}
where $\tau$ is the sampling temperature. PAD, BOS, and UNK tokens are excluded from sampling, while EOS remains available to terminate generation. Decoding stops when EOS is sampled, or the model context limit is reached. The sampling settings are reported in Appendix~\ref{app:evaluation_details}.

\section{Additional experimental details}
\label{app:experimental_details}

\subsection{PROTAC-like pretraining corpus}
\label{app:pretraining_corpus}

The PROTAC-like pretraining corpus described in Section~\ref{sec:experiments} is built using physicochemical and structural filters intended to enrich for large molecules with PROTAC-like properties. We retain compounds with molecular weight of 650--1200~Da, topological polar surface area (TPSA) of 120--320~\AA$^2$, 12--45 rotatable bonds, 8--24 hydrogen-bond acceptors, and 0--8 hydrogen-bond donors.

To favor bifunctional structures, molecules must also contain at least two distinct ring systems. We identify the two largest ring systems and require a minimum topological distance of 8--60 bonds between them.

Molecules satisfying all criteria are converted to canonical isomeric SMILES and deduplicated within each source, yielding 68,550 molecules from ZINC20 and 19,056 from ChEMBL 36. After merging and deduplicating across sources, the final corpus contains 79,406 unique PROTAC-like molecules.

\subsection{POI and E3-ligase annotation harmonization}
\label{app:poi_normalization}

POI and E3-ligase annotations are harmonized across PROTAC-DB, TPDdb, and PROTAC-Pedia before merging the datasets. When a UniProt accession is available, we retrieve the corresponding primary gene name from UniProtKB through its REST API~\citep{ahmad2025uniprot} and use it as the canonical POI label. This ensures consistent target identification across databases despite differences in the original annotations.

When an accession is unavailable, we apply manually curated normalization rules only if the annotation can be mapped unambiguously to a single protein. For example, \texttt{cIAP1} is mapped to \texttt{BIRC2}, \texttt{XIAP} to \texttt{BIRC4}, \texttt{Cereblon} to \texttt{CRBN}, and \texttt{AhR} to \texttt{AHR}. We then use the normalized gene names to retrieve the corresponding protein annotations from UniProtKB, ensuring that these targets follow the same naming convention. Ambiguous annotations remain unresolved to avoid incorrect protein assignments.

Finally, we cross-check the mappings across the three databases to identify inconsistencies and unresolved annotations before constructing the final datasets.

\subsection{Dataset composition and component-level chemistry}
\label{app:dataset_composition}

\begin{figure*}[!t]
\centering
\includegraphics[width=\textwidth]{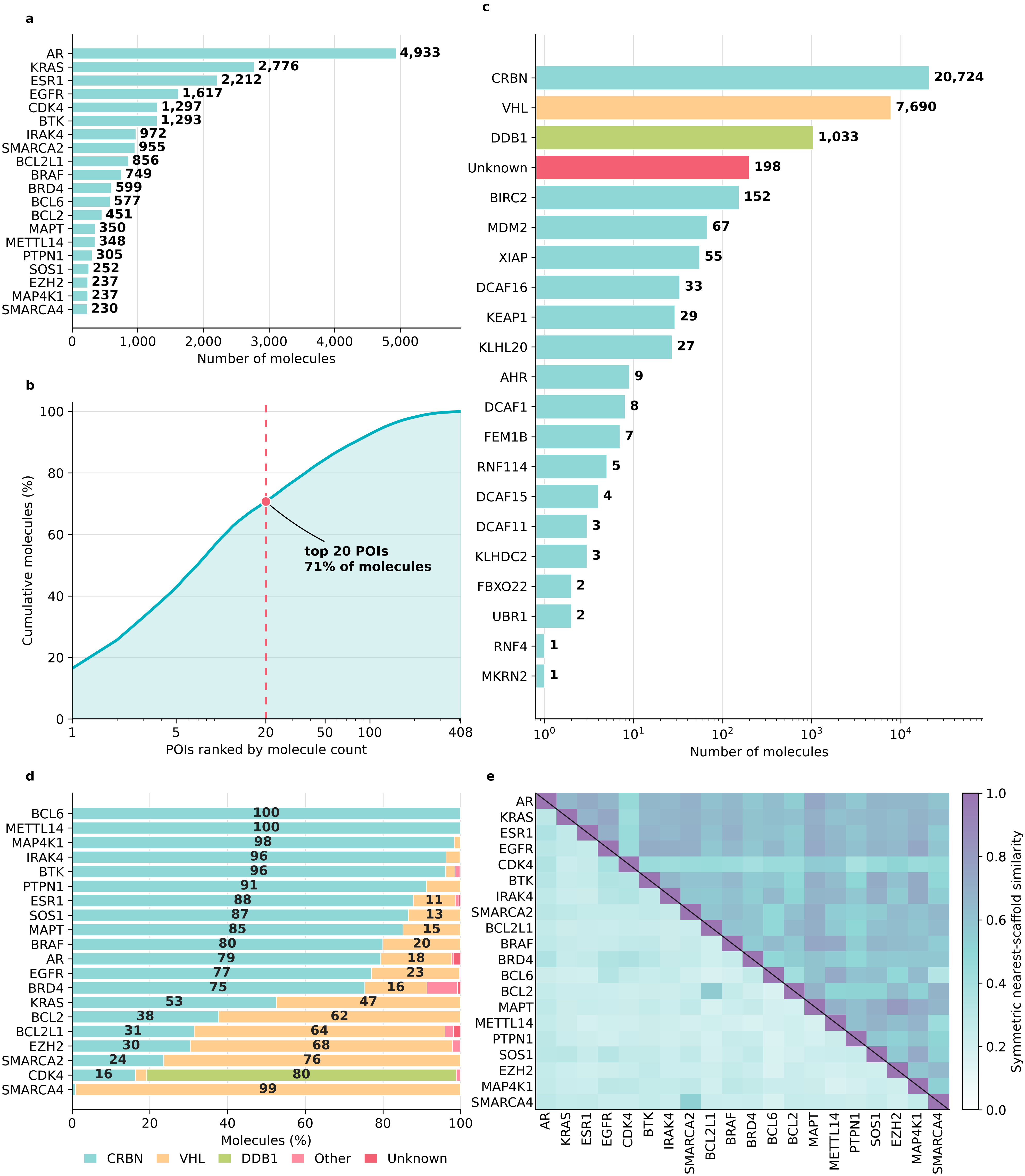}
\caption{
\textbf{Dataset composition and component-level scaffold similarity.}
\textbf{(a)} Number of molecules of seen POIs.
\textbf{(b)} Cumulative fraction of the dataset covered by POIs ranked by molecule count.
\textbf{(c)} E3-ligase distribution across the complete dataset.
\textbf{(d)} E3-ligase composition across the seen POIs.
\textbf{(e)} Pairwise scaffold similarity across the seen POIs, comparing predicted POI-warhead scaffolds (lower triangle) and E3-ligand scaffolds (upper triangle). Similarity is measured using symmetric nearest-scaffold Tanimoto similarity.
}
\label{fig:dataset_overview}
\end{figure*}

Figure~\ref{fig:dataset_overview} shows the composition of the PROTAC dataset, revealing substantial imbalance across POIs. The 20 most represented POIs, referred to as the \emph{seen POIs} in Section~\ref{sec:data}, account for 21,246 molecule--POI pairs (70.7\% of the complete dataset; Figure~\ref{fig:dataset_overview}a,b), with counts ranging from 4,933 for AR to 230 for SMARCA4.

These targets provide sufficient data for separate training, validation, and test subsets. In contrast, many less represented POIs have too few examples for reliable held-out evaluation. We therefore include them only in training to expose the model to a broader range of protein conditions.

E3-ligase usage is also highly imbalanced. CRBN and VHL account for 20,724 and 7,690 molecules, respectively, together representing 94.5\% of the complete dataset (Figure~\ref{fig:dataset_overview}c). Although their relative contributions vary across the seen POIs, many targets share the same E3 ligases (Figure~\ref{fig:dataset_overview}d).

To examine how the chemistry of the two binding components varies across targets, we apply PROTAC-Splitter to molecules targeting the seen POIs. We extract Bemis--Murcko scaffolds from the predicted warheads and E3 ligands, then deduplicate them within each POI.

For two POIs $p$ and $q$, let $\mathcal{S}_p$ and $\mathcal{S}_q$ denote their respective sets of unique component scaffolds. For each scaffold in $p$, we identify its most similar scaffold in $q$ and average these maximum similarities, 
\begin{equation}
D(p\rightarrow q)
=
\frac{1}{|\mathcal{S}_p|}
\sum_{s\in\mathcal{S}_p}
\max_{t\in\mathcal{S}_q}
T(s,t),
\end{equation}
where $T(s,t)$ is the Tanimoto similarity between scaffold fingerprints. Because this measure depends on the comparison direction, we define a symmetric similarity, 
\begin{equation}
S_{\mathrm{sym}}(p,q)
=
\frac{1}{2}
\left[
D(p\rightarrow q)
+
D(q\rightarrow p)
\right].
\end{equation}
We calculate this measure for warhead and E3-ligand scaffolds.

Figure~\ref{fig:dataset_overview}e shows that E3-ligand scaffolds are substantially more similar across POIs than warhead scaffolds, with mean similarities of 0.600 and 0.247, respectively. This indicates that E3-ligand chemistry is more widely shared across targets, whereas warhead chemistry differs more strongly. Consequently, whole-PROTAC similarity may partly reflect shared E3-ligand structures rather than differences related to POI identity. To account for this potential confounding effect, we evaluate chemical similarity both globally and within fixed CRBN and VHL groups (Figure~\ref{fig:poi_signal}).

\subsection{Zero-shot POI selection}
\label{app:zero_shot_selection}

We select four unseen POIs to evaluate zero-shot generalization across different levels of protein similarity. Candidates must fall outside the seen-POI panel and have sufficient reference molecules for held-out evaluation. We also require that their standardized SMILES do not occur under any POI in conditional training. Thus, neither the selected POIs nor their molecules appear in the conditional training. Together, the four zero-shot POIs provide 217 held-out molecule--POI pairs. We measure protein similarity using cosine similarity between mean-pooled ESM-2 embeddings.

We first identify WDR5 and DHFR as the candidate pair most distant from the proteins available for training. For a candidate pair $(a,b)$, let $\mathcal{P}_{\mathrm{train}}^{(a,b)}$ denote the POIs remaining after both candidates are excluded. For each candidate $p\in\{a,b\}$, the distance to its nearest training POI is

\begin{equation}
d_{\mathrm{train}}(p)
=
\min_{q\in\mathcal{P}_{\mathrm{train}}^{(a,b)}}
\left[
1-
\frac{
\mathbf{z}_p^{\top}\mathbf{z}_q
}{
\|\mathbf{z}_p\|_2\|\mathbf{z}_q\|_2
}
\right],
\end{equation}

where $\mathbf{z}_p$ and $\mathbf{z}_q$ are mean-pooled ESM-2 embeddings. Larger distances correspond to lower similarity to the nearest training POI.

To ensure that both candidates are distant from training, we score each pair using the smaller of the two distances,

\begin{equation}
D_{\mathrm{pair}}(a,b)
=
\min
\left\{
d_{\mathrm{train}}(a),
d_{\mathrm{train}}(b)
\right\}.
\end{equation}

We select the pair maximizing this score. WDR5 and DHFR have nearest-training distances of $0.1234$ and $0.0864$, respectively.

We then select two additional zero-shot targets to cover higher levels of protein similarity. For each remaining candidate $p$, we calculate its maximum cosine similarity to the seen POIs,

\begin{equation}
S_{20}(p)
=
\max_{q\in\mathcal{P}_{20}}
\frac{
\mathbf{z}_p^{\top}\mathbf{z}_q
}{
\|\mathbf{z}_p\|_2\|\mathbf{z}_q\|_2
},
\end{equation}

where $\mathcal{P}_{20}$ denotes the set of seen POIs. FGFR3 lies at the median of the similarity distribution and is selected as the medium-similarity target. MAP3K13 has the highest similarity and is selected as the high-similarity target.

Table~\ref{tab:zero_shot_selection} reports each selected POI and its nearest match in the seen-POI panel. WDR5 and DHFR represent lower-similarity conditions, while FGFR3 and MAP3K13 provide medium- and high-similarity conditions. This selection allows us to examine whether zero-shot performance varies with similarity in ESM-2 embedding space.

\begin{table}[H]
\centering
\small
\caption{
\textbf{Protein-representation similarity of the four zero-shot POIs.}
For each zero-shot POI, the closest seen POI is identified by cosine similarity between mean-pooled ESM-2 embeddings.
}
\label{tab:zero_shot_selection}
\setlength{\tabcolsep}{5pt}
\begin{tabular}{llrll}
\hline
\textbf{Role} &
\textbf{POI} &
\textbf{Records} &
\textbf{Closest top-20 POI} &
\textbf{Cosine similarity} \\
\hline
Most dissimilar & WDR5    & 66 & PTPN1  & 0.760 \\
Dissimilar      & DHFR    & 43 & EGFR   & 0.894 \\
Medium          & FGFR3   & 58 & MAP4K1 & 0.969 \\
Very similar    & MAP3K13 & 50 & AR     & 0.987 \\
\hline
\end{tabular}
\end{table}

\subsection{Conditional data splitting}
\label{app:data_splits}
After excluding the four zero-shot POIs, 29,836 molecule--POI pairs across 404 POIs remain for conditional model development. The seen POIs are AR, KRAS, ESR1, EGFR, CDK4, BTK, IRAK4, SMARCA2, BCL2L1, BRAF, BRD4, BCL6, BCL2, MAPT, METTL14, PTPN1, SOS1, EZH2, MAP4K1, and SMARCA4. Together, they account for 21,246 pairs (71.2\% of the remaining dataset).

To prevent molecular leakage between partitions, we group all records sharing the same standardized SMILES before splitting. Each group is assigned entirely to training, validation, or test, ensuring that the same molecular structure never appears in different partitions.

We split the records of seen POIs into 80\% training, 10\% validation, and 10\% test subsets. The resulting top-20 split contains 16,994 training, 2,126 validation, and 2,126 test pairs. The 384 training-only POIs contribute another 8,590 training pairs, yielding 25,584 training, 2,126 validation, and 2,126 test pairs overall.

The exact target-wise counts are reported in Table~\ref{tab:conditional_split_top20}.

\begin{table}[t]
\centering
\small
\caption{
Target-wise counts for the seen POIs. Data for each POI are split into 80\% training, 10\% validation, and 10\% test subsets while keeping molecular structures disjoint across partitions.
}
\label{tab:conditional_split_top20}
\setlength{\tabcolsep}{6pt}
\begin{tabular}{lrrrr}
\hline
\textbf{POI} & \textbf{Total} & \textbf{Train} & \textbf{Val} & \textbf{Test} \\
\hline
AR       & 4933 & 3947 & 493 & 493 \\
KRAS     & 2776 & 2220 & 278 & 278 \\
ESR1     & 2212 & 1770 & 221 & 221 \\
EGFR     & 1617 & 1293 & 162 & 162 \\
CDK4     & 1297 & 1037 & 130 & 130 \\
BTK      & 1293 & 1035 & 129 & 129 \\
IRAK4    & 972  & 778  & 97  & 97  \\
SMARCA2  & 955  & 763  & 96  & 96  \\
BCL2L1   & 856  & 684  & 86  & 86  \\
BRAF     & 749  & 599  & 75  & 75  \\
BRD4     & 599  & 479  & 60  & 60  \\
BCL6     & 577  & 461  & 58  & 58  \\
BCL2     & 451  & 361  & 45  & 45  \\
MAPT     & 350  & 280  & 35  & 35  \\
METTL14  & 348  & 278  & 35  & 35  \\
PTPN1    & 305  & 245  & 30  & 30  \\
SOS1     & 252  & 202  & 25  & 25  \\
EZH2     & 237  & 189  & 24  & 24  \\
MAP4K1   & 237  & 189  & 24  & 24  \\
SMARCA4  & 230  & 184  & 23  & 23  \\
\hline
\textbf{Total} & \textbf{21246} & \textbf{16994} & \textbf{2126} & \textbf{2126} \\
\hline
\end{tabular}
\end{table}

\subsection{POI-associated chemical-space characterization}
\label{app:poi_structure}

We examine whether PROTAC chemistry in the reference dataset exhibits patterns related to POI identity. All analyses use 2048-bit Morgan fingerprints with radius 2, and chemical similarity is measured using the Tanimoto coefficient,
\begin{equation}
T(\mathbf{f}_i,\mathbf{f}_j)
=
\frac{
\mathbf{f}_i^\top \mathbf{f}_j
}{
\|\mathbf{f}_i\|_1
+
\|\mathbf{f}_j\|_1
-
\mathbf{f}_i^\top \mathbf{f}_j
}.
\end{equation}

\paragraph{Chemical-space visualization.}

We sample 300 molecules from each of the 10 most represented POIs, yielding a balanced set of 3,000 molecules. Equal sampling prevents targets with more examples from dominating the visualization. We first reduce the fingerprint matrix to 50 principal components using PCA, then apply t-SNE to obtain a two-dimensional embedding.

Figure~\ref{fig:poi_signal}a,b shows the same coordinates colored by E3-ligase and POI identity, respectively. Broad regions reflect E3 identity, while smaller regions are enriched for individual POIs. Several POIs occupy separated regions, that their molecules span different chemotypes.

\paragraph{Global POI-associated similarity.}

We assess whether molecules targeting the same POI are more chemically similar than those targeting different POIs. Because the seen POIs have unequal numbers of records, sampling molecular pairs directly from the pooled dataset would favor targets with more examples, such as AR. We therefore balance pair sampling across POIs.

For same-POI comparisons, we uniformly sample a seen POI and select two distinct molecules targeting it. For different-POI comparisons, we uniformly sample two distinct POIs and select one molecule from each. We repeat this procedure to obtain 10,000 pairs of each type. A molecule may be reused across comparisons, but the two molecules in each same-POI pair are distinct.

Let $\mathcal{P}_{\mathrm{same}}$ and $\mathcal{P}_{\mathrm{diff}}$ denote the sampled pair sets, each containing $M=10{,}000$ pairs. Their mean Tanimoto similarities are
\begin{equation}
\overline{T}_{\mathrm{same}}
=
\frac{1}{M}
\sum_{(i,j)\in\mathcal{P}_{\mathrm{same}}}
T(\mathbf{f}_i,\mathbf{f}_j),
\qquad
\overline{T}_{\mathrm{diff}}
=
\frac{1}{M}
\sum_{(i,j)\in\mathcal{P}_{\mathrm{diff}}}
T(\mathbf{f}_i,\mathbf{f}_j),
\end{equation}
and their difference is
\begin{equation}
\Delta_T
=
\overline{T}_{\mathrm{same}}
-
\overline{T}_{\mathrm{diff}}.
\end{equation}

Mean similarity is 0.385 for same-POI pairs and 0.190 for different-POI pairs, giving $\Delta_T=0.195$. This difference shows that molecules targeting the same target are more similar on average. However, the moderate same-POI similarity also suggests that a single target can span multiple chemotypes rather than forming a single compact chemical cluster.

\paragraph{Controlling for E3-ligase identity.}

Because E3-binding ligands contribute to whole-PROTAC similarity, part of the observed signal may reflect shared E3 chemistry. To account for this, we repeat the pairwise analysis within CRBN and VHL groups. Within each group, we uniformly sample POIs to construct 10,000 same-POI and 10,000 different-POI pairs.

Same-POI similarity remains higher in both E3 groups:
\begin{equation}
\begin{aligned}
\mathrm{CRBN}: \quad&
\overline{T}_{\mathrm{same}}=0.421,
&
\overline{T}_{\mathrm{diff}}=0.223,\\
\mathrm{VHL}: \quad&
\overline{T}_{\mathrm{same}}=0.535,
&
\overline{T}_{\mathrm{diff}}=0.329.
\end{aligned}
\end{equation}

The similarity gaps are 0.198 for CRBN and 0.206 for VHL, comparable to the global gap of 0.195. Thus, POI-specific chemical similarity persists when E3 identity is fixed and cannot be explained solely by differences between CRBN- and VHL-based chemistry.

\paragraph{Local POI organization.}

Global pairwise similarity does not imply that all molecules targeting the same POI form a single cluster. A target may instead be represented by several chemotypes, each of which is locally enriched for that POI. We therefore assess this local organization using nearest-neighbor POI purity.

The analysis uses molecules for the seen POIs within each CRBN or VHL group. For query molecule $i$ with POI label $y_i$, let $\mathcal{N}^{(e)}_k(i)$ denote its $k$ nearest molecules by Tanimoto similarity within E3 group $e$, excluding the query itself. We consider neighborhood sizes $k=1,\ldots,30$.

For each query, POI purity is defined as the fraction of neighbors sharing its POI label,
\begin{equation}
\rho_k^{(e)}(i)
=
\frac{1}{k}
\sum_{j\in\mathcal{N}^{(e)}_k(i)}
\mathbf{1}
\left[
y_j=y_i
\right].
\end{equation}

For an E3 group containing $N_e$ query molecules, mean purity is
\begin{equation}
\overline{\rho}_k^{(e)}
=
\frac{1}{N_e}
\sum_{i=1}^{N_e}
\rho_k^{(e)}(i).
\end{equation}

\paragraph{Within-E3 shuffled-label null.}

Unequal POI frequencies can increase same-POI neighborhood matches by chance. To account for this, we construct a shuffled-label null separately within each E3 group. Molecular fingerprints and nearest-neighbor relationships remain fixed, while POI labels are randomly permuted. This preserves target frequencies but breaks the relationship between POI identity and chemical structure.

Let $y_i^{(b)}$ denote the label assigned to molecule $i$ in permutation $b$. The corresponding query-level null purity is
\begin{equation}
\rho_{k,\mathrm{null}}^{(e,b)}(i)
=
\frac{1}{k}
\sum_{j\in\mathcal{N}^{(e)}_k(i)}
\mathbf{1}
\left[
y_j^{(b)}=y_i^{(b)}
\right],
\end{equation}
and its mean across the E3 group is
\begin{equation}
\overline{\rho}_{k,\mathrm{null}}^{(e,b)}
=
\frac{1}{N_e}
\sum_{i=1}^{N_e}
\rho_{k,\mathrm{null}}^{(e,b)}(i).
\end{equation}

We repeat the permutation procedure $B=300$ times per E3 group. The mean null purity is
\begin{equation}
\overline{\rho}_{k,\mathrm{null}}^{(e)}
=
\frac{1}{B}
\sum_{b=1}^{B}
\overline{\rho}_{k,\mathrm{null}}^{(e,b)},
\end{equation}
with the 2.5th--97.5th percentile range across permutations shown in Figure~\ref{fig:poi_signal}d.

Observed purity remains substantially higher than the shuffled-label null across neighborhood sizes. At $k=30$, purity reaches 0.977 for CRBN and 0.933 for VHL, compared with null expectations of 0.124 and 0.137, respectively. This indicates that local enrichment for POI identity cannot be explained by unequal target frequencies.

\paragraph{Interpretation.}

Two complementary patterns emerge. Globally, molecules targeting the same POI are more similar than those targeting different POIs, even within fixed E3 groups. However, the moderate mean same-POI similarity indicates that individual targets can span multiple chemotypes.

Locally, molecules with similar chemical structures are strongly enriched for the same POI, well above the shuffled-label null. PROTAC chemical space is therefore \emph{globally heterogeneous but locally organized} with respect to POI identity.

These findings motivate complementary fidelity measures: likelihood preference under the requested protein condition and POI enrichment among held-out reference neighbors. Neither measure assumes one chemical centroid per POI.

\subsection{Generation and metric implementation}
\label{app:evaluation_details}

The same decoding settings are used for the final unconditional and conditional generation experiments. Generation begins with the BOS token and uses temperature $\tau=1.0$ and nucleus sampling with $p=0.9$, without top-$k$ filtering. We repeat generation using seeds 71, 101, and 123. PAD, BOS, and UNK tokens are excluded from sampling, while EOS can terminate generation. The maximum sequence length is 320 tokens, allowing up to 319 new tokens after BOS.

Generated SMILES are parsed and sanitized using RDKit. Molecules that pass both steps are considered valid and converted to canonical isomeric SMILES for deduplication.

For conditional evaluation, deduplication is performed independently for each POI. Repeated molecules count only once within a given target but can be counted separately if generated under different POI conditions. Novelty is evaluated against the entire conditional training set, regardless of target identity. A molecule is therefore considered non-novel if its canonical structure appears anywhere in the training data.

Internal diversity is calculated over unique valid molecules using 2048-bit Morgan fingerprints with radius 2. When more than 1,000 molecules are available, a random subset of 1,000 is selected for pairwise Tanimoto similarity calculations.

Physicochemical descriptors are also computed over unique valid molecules, including molecular weight, TPSA, LogP, heavy-atom count, rotatable-bond count, aromatic and non-aromatic ring counts, and hydrogen-bond acceptor and donor counts.

\subsection{Physicochemical comparison of PROTAC-like, PROTAC, and generated molecules}
\label{app:physicochemical_comparison}

\begin{figure*}[!htbp]
\centering
\includegraphics[width=\textwidth]{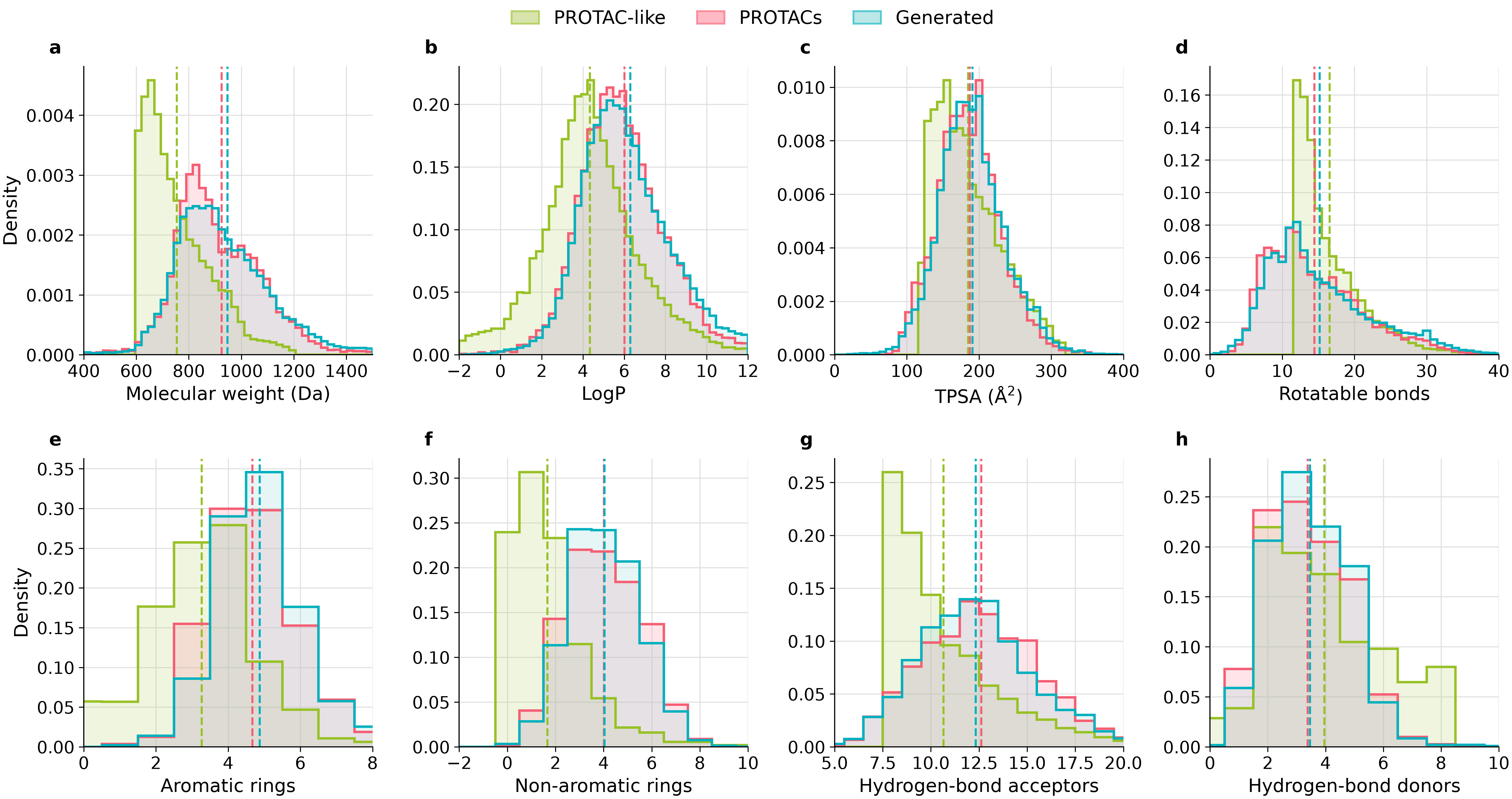}
\caption{
\textbf{Physicochemical distributions of PROTAC-like, reference PROTAC, and generated molecules.}
Normalized density distributions for PROTAC-like structures, reference PROTACs, and molecules generated by the conditional model. Dashed vertical lines indicate distribution means.
\textbf{(a)} Molecular weight (MW).
\textbf{(b)} LogP.
\textbf{(c)} Topological polar surface area (TPSA).
\textbf{(d)} Number of rotatable bonds.
\textbf{(e)} Number of aromatic rings.
\textbf{(f)} Number of non-aromatic rings.
\textbf{(g)} Number of hydrogen-bond acceptors (HBA).
\textbf{(h)} Number of hydrogen-bond donors (HBD).
}
\label{fig:physicochemical_comparison}
\end{figure*}

Figure~\ref{fig:physicochemical_comparison} compares physicochemical distributions across the PROTAC-like pretraining corpus, reference PROTACs, and conditionally generated molecules. Each property uses shared histogram bins and normalized densities across datasets. Dashed vertical lines indicate the mean of each distribution.

The PROTAC-like corpus broadly overlaps the reference PROTAC distributions, with differences consistent with its property-based construction from general molecular databases. Generated molecules also overlap the reference distributions across these descriptors. This comparison complements the generation-quality metrics in Section~\ref{sec:evaluation}.

\subsection{Conditional-fidelity evaluation}
\label{app:conditional_fidelity}

Conditional fidelity is evaluated using the held-out test set.
The reference pool consists of unique test molecules, each with
its corresponding POI labels. For a reference molecule $x$,
we define the set of POIs it targets as

\begin{equation}
\mathcal{P}(x)
=
\left\{
p \in \mathcal{T}
:
x \text{ is annotated as targeting } p
\text{ in the held-out test set}
\right\},
\end{equation}

where $\mathcal{T}$ is the set of POIs included in the evaluation.

For example, suppose the evaluation includes $\mathcal{T}=\{\mathrm{AR},\mathrm{BTK},\mathrm{EGFR}\}$.
 If a held-out molecule $x$ targets both AR and BTK, then
$\mathcal{P}(x)=\{\mathrm{AR},\mathrm{BTK}\}$.
The molecule appears once in the reference pool, but both AR and BTK are treated as correct conditions.

\paragraph{Likelihood-based conditional fidelity.}

Each reference molecule is evaluated under every POI condition
in $\mathcal{T}$ by computing its negative log-likelihood ($\mathrm{NLL}$).

For each POI $p \in \mathcal{P}(x)$, the pair $(x,p)$ defines one correct-condition evaluation. The competing POIs are defined as
\begin{equation}
\mathcal{C}(x)
=
\mathcal{T}
\setminus
\mathcal{P}(x).
\end{equation}

This excludes all known targets of $x$ from the competing set, ensuring that a correct POI is never treated as an incorrect condition.

For each correct molecule--POI pair $(x,p)$, conditional preference is calculated as
\begin{equation}
\Delta_{\mathrm{NLL}}(x,p)
=
\frac{1}{|\mathcal{C}(x)|}
\sum_{q\in\mathcal{C}(x)}
\mathrm{NLL}(x\mid q)
-
\mathrm{NLL}(x\mid p).
\end{equation}

A positive value indicates lower $\mathrm{NLL}$ under the correct condition than the mean across competing conditions. For a multi-target molecule, we calculate the score separately for each correct POI. If $\mathcal{C}(x)$ is empty, the molecule is excluded because there is no competing condition.

In the previous example, $\mathcal{P}(x)=\{\mathrm{AR},\mathrm{BTK}\}$, so $\mathcal{C}(x)=\{\mathrm{EGFR}\}$. The molecule contributes two evaluations, $(x,\mathrm{AR})$ and $(x,\mathrm{BTK})$, with EGFR as the only competing condition. Neither AR nor BTK is treated as incorrect when evaluating the other.

\paragraph{Nearest-neighbor query construction.}

Nearest-neighbor fidelity uses valid, unique, training-novel generated molecules. A generated molecule may be absent from training yet identical to a held-out test molecule. We retain such queries because they represent recovery of reference chemistry not seen during fine-tuning.

Generated queries and reference molecules are encoded using 2048-bit Morgan fingerprints with radius 2. For each molecule $g$ generated under POI $p$, we calculate its Tanimoto similarity to every molecule in the reference pool. Its $k$ nearest reference molecules form the nearest-neighbor set $\mathcal{N}_k(g)$, where $k\in\{5,10,20\}$.

\paragraph{Multi-target handling in nearest-neighbor purity.}

For a molecule $g$ generated under condition $p$, a reference neighbor $r$ contributes to purity whenever $p$ is among the POIs targeted by $r$:
\begin{equation}
p \in \mathcal{P}(r).
\end{equation}

This allows a reference molecule targeting multiple POIs to count as a correct neighbor for any of those targets. POI purity is defined as
\begin{equation}
\rho_k(g,p)
=
\frac{1}{k}
\sum_{r\in\mathcal{N}_k(g)}
\mathbf{1}
\left[
p\in\mathcal{P}(r)
\right],
\end{equation}
as defined in Section~\ref{sec:evaluation}. The indicator equals 1 if reference molecule $r$ targets POI $p$ and 0 otherwise. Purity is calculated for each generated query and then averaged across queries for the same POI.

For example, the reference molecule $x$ targeting AR and BTK contributes to purity for queries generated under either AR or BTK. It does not contribute to purity for EGFR queries because $\mathrm{EGFR}\notin\mathcal{P}(x)$.

\subsection{POI-warhead processing}
\label{app:warhead_details}

POI-warhead evaluation is performed using unique valid PROTACs generated under each POI condition.  We apply PROTAC-Splitter to decompose these molecules into their structural components for subsequent warhead analysis.

\paragraph{Warhead extraction.}

PROTAC-Splitter returns three dot-separated SMILES components in the following order:
\begin{equation}
\text{E3 ligand}
\;.\;
\text{linker}
\;.\;
\text{POI warhead}.
\end{equation}
We retain predictions with exactly three components and identify the third as the POI-binding warhead. Predictions with missing or additional components are excluded because the warhead cannot be reliably identified.

\paragraph{Parent-warhead construction.}
The extracted warhead may contain an attachment marker, \texttt{[\string*:1]}, at its linker connection. This marker identifies the split point but is not part of the standalone warhead used for evaluation. We replace it with hydrogen to obtain a parent warhead for chemical comparison.

We parse each predicted fragment with RDKit. If an attachment marker is present, its dummy atom must have exactly one neighbor connected by a single bond. We replace the dummy atom with hydrogen and sanitize the molecule. Explicit hydrogens are then removed where chemically appropriate, followed by a second sanitization. We exclude fragments if conversion or sanitization fails or if any dummy atom remains.

We convert successfully processed parent warheads to canonical isomeric SMILES and apply the same procedure to training and generated PROTACs.

\paragraph{Target-specific training reference.}

For each POI $p$, we extract warheads from conditional training PROTACs targeting $p$. We apply the same parent-warhead construction procedure and remove duplicate structures. The resulting set of unique canonical parent warheads forms the target-specific training reference $\mathcal{R}^{W}_p$.

A generated warhead is considered non-novel for POI $p$ only if its canonical structure appears in $\mathcal{R}^{W}_p$. Its occurrence under another POI does not affect novelty for the current target.

Warhead uniqueness, novelty, and internal diversity are defined in Section~\ref{sec:evaluation}. Internal diversity is calculated over unique successfully processed generated warheads using 2048-bit Morgan fingerprints with radius 2.

\paragraph{Processing coverage.}

Warhead evaluation requires two successful processing steps. First, PROTAC-Splitter must produce a usable three-component decomposition. Second, the extracted POI warhead must be converted into a valid parent structure. We measure the coverage of each step.

For POI $p$, let $N^{G}_{p}$ denote the number of unique valid generated PROTACs, $N^{S}_{p}$ the number with a usable three-component decomposition, and $N^{W}_{p}$ the number successfully converted to parent warheads. We define the coverage metrics as
\begin{equation}
\mathrm{SplitCov}(p)
=
\frac{N^{S}_{p}}{N^{G}_{p}},
\qquad
\mathrm{WarheadCov}(p)
=
\frac{N^{W}_{p}}{N^{S}_{p}}.
\end{equation}

$\mathrm{SplitCov}$ is the proportion of generated PROTACs successfully decomposed, and $\mathrm{WarheadCov}$ is the proportion of those decompositions yielding valid parent warheads.

Target-wise coverage results are reported in Appendix Table~\ref{tab:warhead_coverage}.

\section{Additional results}
\label{app:additional_results}

\subsection{POI-warhead generation analysis}
\label{app:warhead_results}

Table~\ref{tab:warhead_results} reports warhead-level generation metrics for the seen POIs, complementing the results in Section~\ref{sec:seen_poi_results}. Warheads are extracted using PROTAC-Splitter and processed as described in Appendix~\ref{app:warhead_details}. Metrics are evaluated across three generation seeds.

Warhead diversity varies across POIs. BRD4 and MAP4K1 have the highest internal diversity, while METTL14 and BCL2 have the lowest.

Molecular uniqueness also differs from warhead uniqueness. For example, BCL6 achieves $28.50\%$ molecular uniqueness but only $8.61\%$ warhead uniqueness, indicating that distinct PROTACs can share the same POI-binding warhead.

Warhead novelty does not establish target relevance. SMARCA4, for example, achieves $95.87\%$ warhead novelty but only $27.37\%$ generated 5-NN purity. 

\begin{table}[t]
\caption{
\textbf{Warhead-level generation metrics across the seen POIs.}
Warheads are identified using PROTAC-Splitter and converted to parent structures. Uniqueness is measured over successfully processed parent warheads, and novelty is evaluated against parent warheads extracted from training PROTACs for the same POI. Values are reported as mean (sample standard deviation) across three seeds. Best per-target values in each column are shown in bold.
}
\label{tab:warhead_results}
\centering
\scriptsize
\setlength{\tabcolsep}{5pt}
\resizebox{\linewidth}{!}{%
\begin{tabular}{lccccc}
\hline
\textbf{POI} &
\textbf{Unique PROTACs} &
\textbf{Unique warheads} &
\textbf{Warhead uniqueness (\%)} &
\textbf{Warhead novelty (\%)} &
\textbf{Warhead diversity} \\
\hline
AR      & 4,847 (50) & 1,005 (30) & 20.75 (0.40) & 61.20 (0.39) & 0.811 (0.001) \\
ESR1    & 4,093 (37) & 893 (33) & 21.85 (0.64) & 69.09 (0.50) & 0.776 (0.005) \\
CDK4    & 3,047 (13) & 619 (15) & 20.31 (0.52) & 72.99 (1.88) & 0.780 (0.004) \\
KRAS    & 4,954 (25) & 898 (29) & 18.12 (0.62) & 80.01 (1.10) & 0.749 (0.005) \\
EZH2    & 3,100 (54) & 788 (41) & 25.42 (1.13) & 96.14 (0.23) & 0.758 (0.005) \\
BTK     & 4,214 (24) & 1,350 (50) & 32.04 (1.05) & 84.86 (0.53) & 0.817 (0.001) \\
BRD4    & 4,045 (32) & 1,125 (25) & 27.86 (0.46) & 94.13 (0.19) & \textbf{0.869 (0.001)} \\
BRAF    & 2,869 (24) & 450 (22) & 15.70 (0.71) & 87.49 (0.81) & 0.756 (0.003) \\
SMARCA4 & 4,061 (38) & 1,291 (45) & 31.80 (0.88) & 95.87 (0.13) & 0.766 (0.006) \\
MAP4K1  & 3,332 (21) & 1,163 (17) & 34.93 (0.33) & \textbf{96.41 (0.18)} & 0.835 (0.003) \\
IRAK4   & \textbf{5,580 (63)} & \textbf{1,588 (26)} & 28.47 (0.27) & 90.54 (0.55) & 0.816 (0.002) \\
EGFR    & 5,096 (38) & 1,384 (11) & 27.17 (0.29) & 82.85 (0.48) & 0.803 (0.002) \\
BCL2L1  & 3,685 (25) & 784 (36) & 21.28 (0.94) & 85.09 (1.09) & 0.762 (0.006) \\
SMARCA2 & 4,152 (44) & 1,352 (34) & 32.56 (0.51) & 87.66 (0.62) & 0.776 (0.001) \\
METTL14 & 1,372 (28) & 156 (5) & 11.39 (0.22) & 86.13 (0.68) & 0.581 (0.003) \\
PTPN1   & 2,401 (23) & 491 (25) & 20.48 (0.98) & 93.39 (0.62) & 0.760 (0.007) \\
BCL2    & 2,809 (42) & 1,417 (15) & \textbf{50.47 (0.55)} & 95.32 (0.07) & 0.633 (0.001) \\
SOS1    & 3,374 (60) & 996 (48) & 29.51 (0.91) & 95.64 (0.21) & 0.786 (0.006) \\
BCL6    & 2,830 (64) & 244 (4) & 8.61 (0.20) & 85.91 (0.94) & 0.791 (0.006) \\
MAPT    & 3,004 (17) & 594 (16) & 19.78 (0.56) & 91.75 (0.38) & 0.729 (0.005) \\
\hline
\textbf{Macro} &
\textbf{3,643 (3)} &
\textbf{929 (8)} &
\textbf{24.93 (0.20)} &
\textbf{86.62 (0.15)} &
\textbf{0.768 (0.001)} \\
\hline
\end{tabular}%
}
\end{table}

\paragraph{Warhead coverage.}

PROTAC-Splitter produces usable three-component decompositions for nearly all unique valid generated PROTACs. Across the seen POIs, macro splitter coverage reaches $99.99\%$, and $99.97\%$ of the extracted warheads are successfully converted to valid parent structures. Table~\ref{tab:warhead_coverage} reports the target-wise results.

\begin{table}[t]
\caption{
\textbf{PROTAC-Splitter and warhead processing coverage for seen POIs.}
Splitter coverage measures the proportion of unique valid generated PROTACs with a usable three-component decomposition. Warhead processing coverage measures the proportion of extracted POI warheads successfully converted to valid parent structures. Values are reported as mean (sample standard deviation) across three generation seeds. The macro row reports the unweighted mean across POIs.
}
\label{tab:warhead_coverage}
\label{tab:splitter_coverage}
\centering
\scriptsize
\setlength{\tabcolsep}{4pt}
\begin{tabular}{lcc}
\hline
\textbf{POI} &
\textbf{Splitter coverage (\%)} &
\textbf{Warhead processing (\%)} \\
\hline
AR      & \textbf{100.00 (0.00)} & 99.91 (0.03) \\
ESR1    & 99.99 (0.01) & 99.88 (0.07) \\
CDK4    & \textbf{100.00 (0.00)} & \textbf{100.00 (0.00)} \\
KRAS    & \textbf{100.00 (0.00)} & \textbf{100.00 (0.00)} \\
EZH2    & \textbf{100.00 (0.00)} & 99.96 (0.02) \\
BTK     & \textbf{100.00 (0.00)} & \textbf{100.00 (0.00)} \\
BRD4    & 99.99 (0.01) & 99.88 (0.04) \\
BRAF    & 99.99 (0.02) & \textbf{100.00 (0.00)} \\
SMARCA4 & 99.98 (0.03) & \textbf{100.00 (0.00)} \\
MAP4K1  & 99.92 (0.02) & 99.97 (0.00) \\
IRAK4   & \textbf{100.00 (0.00)} & 99.97 (0.01) \\
EGFR    & \textbf{100.00 (0.00)} & 99.97 (0.01) \\
BCL2L1  & 99.99 (0.02) & 99.99 (0.02) \\
SMARCA2 & 99.99 (0.01) & 99.99 (0.01) \\
METTL14 & \textbf{100.00 (0.00)} & \textbf{100.00 (0.00)} \\
PTPN1   & 99.97 (0.05) & 99.87 (0.04) \\
BCL2    & 99.99 (0.02) & 99.98 (0.02) \\
SOS1    & \textbf{100.00 (0.00)} & 99.98 (0.02) \\
BCL6    & \textbf{100.00 (0.00)} & \textbf{100.00 (0.00)} \\
MAPT    & 99.97 (0.00) & 99.99 (0.02) \\
\hline
\textbf{Macro} & \textbf{99.99 (0.00)} & \textbf{99.97 (0.00)} \\
\hline
\end{tabular}
\end{table}

\subsection{POI-neighborhood purity: target-wise reference, conditional, and unconditional models}
\label{app:knn_calibration}
We compare the $k$-NN purity of held-out reference molecules with that of conditional and unconditional generation. Reference purity is calculated using leave-one-out nearest neighbors within the test set, while both generated molecular sets are evaluated against the same held-out reference pool. This comparison provides context for interpreting generated neighborhood enrichment relative to the local POI organization of the reference chemistry.

\paragraph{Seen POIs.}
Table~\ref{tab:seen_real_gen_knn} reports the target-wise reference, conditional, and unconditional $k$-NN purity values across the seen POIs. Macro 5-NN purity reaches $92.27\%$ for the reference molecules, $84.73\%$ for TACFormer, and $5.02\%$ for unconditional generation. Conditional purity exceeds unconditional purity for every target at $k\in\{5,10,20\}$, confirming consistent improvements in POI-specific neighborhood enrichment.

The reference--generation gap is largest for SMARCA4 at $k=5$, whereas METTL14 and BCL6 show closely matched purity. These differences indicate that the degree of local POI enrichment varies across the conditional output distributions.

\begingroup
\small
\setlength{\tabcolsep}{10pt}
\setlength{\LTleft}{\fill}
\setlength{\LTright}{\fill}

\begin{longtable}{llccc}

\caption{
\textbf{Reference, conditional, and unconditional $k$-NN purity for seen POIs.}
Reference purity is fixed across seeds. TACFormer and unconditional results are mean (sample standard deviation) across three seeds. All values are percentages; macro values are unweighted means across seen POIs. Bold values indicate the highest target-wise results within each evaluation type and neighborhood size.
}
\label{tab:seen_real_gen_knn} \\

\hline
\textbf{POI} &
\textbf{Evaluation} &
\textbf{5-NN} &
\textbf{10-NN} &
\textbf{20-NN} \\
\hline
\endfirsthead

\multicolumn{5}{l}{\textit{Table \thetable\ (continued)}} \\
\hline
\textbf{POI} &
\textbf{Evaluation} &
\textbf{5-NN} &
\textbf{10-NN} &
\textbf{20-NN} \\
\hline
\endhead

\hline
\multicolumn{5}{r}{\footnotesize\textit{Continued on next page}} \\
\endfoot

\endlastfoot

\multirow[t]{3}{*}{AR}
& Reference & 98.50 & 98.05 & 96.68 \\*
& TACFormer & 97.89 (0.15) & 97.32 (0.16) & 96.33 (0.15) \\*
& Unconditional & \textbf{22.75 (0.14)} & \textbf{23.18 (0.13)} & \textbf{24.01 (0.15)} \\
\hline

\multirow[t]{3}{*}{ESR1}
& Reference & 97.74 & 97.42 & 94.89 \\*
& TACFormer & 97.94 (0.20) & 97.12 (0.25) & 94.78 (0.23) \\*
& Unconditional & 9.91 (0.02) & 10.27 (0.02) & 10.68 (0.03) \\
\hline

\multirow[t]{3}{*}{CDK4}
& Reference & 98.31 & 96.69 & 95.19 \\*
& TACFormer & 95.17 (0.10) & 93.62 (0.23) & 92.66 (0.36) \\*
& Unconditional & 5.80 (0.08) & 5.68 (0.09) & 5.65 (0.09) \\
\hline

\multirow[t]{3}{*}{KRAS}
& Reference & 98.35 & 96.26 & 92.79 \\*
& TACFormer & 97.91 (0.08) & 97.18 (0.06) & 95.56 (0.09) \\*
& Unconditional & 9.96 (0.02) & 10.04 (0.04) & 10.15 (0.04) \\
\hline

\multirow[t]{3}{*}{EZH2}
& Reference & 95.83 & 90.83 & 70.42 \\*
& TACFormer & 82.38 (0.25) & 74.63 (0.13) & 56.15 (0.30) \\*
& Unconditional & 1.56 (0.03) & 1.62 (0.03) & 1.51 (0.02) \\
\hline

\multirow[t]{3}{*}{BTK}
& Reference & 92.71 & 90.08 & 85.54 \\*
& TACFormer & 88.58 (0.21) & 85.50 (0.20) & 81.07 (0.21) \\*
& Unconditional & 6.26 (0.02) & 6.27 (0.02) & 6.32 (0.01) \\
\hline

\multirow[t]{3}{*}{BRD4}
& Reference & 81.00 & 72.33 & 57.08 \\*
& TACFormer & 66.19 (0.77) & 59.97 (0.82) & 50.83 (0.73) \\*
& Unconditional & 5.32 (0.04) & 5.03 (0.03) & 4.59 (0.01) \\
\hline

\multirow[t]{3}{*}{BRAF}
& Reference & 98.13 & 95.60 & 83.40 \\*
& TACFormer & 92.59 (0.27) & 90.22 (0.33) & 80.50 (0.32) \\*
& Unconditional & 2.94 (0.08) & 3.06 (0.07) & 3.09 (0.05) \\
\hline

\multirow[t]{3}{*}{SMARCA4}
& Reference & 77.39 & 73.48 & 60.22 \\*
& TACFormer & 27.37 (1.20) & 26.29 (1.17) & 23.59 (0.98) \\*
& Unconditional & 0.93 (0.02) & 0.91 (0.02) & 0.89 (0.02) \\
\hline

\multirow[t]{3}{*}{MAP4K1}
& Reference & 70.00 & 43.33 & 23.33 \\*
& TACFormer & 61.59 (0.74) & 40.43 (0.49) & 22.65 (0.15) \\*
& Unconditional & 1.54 (0.02) & 1.27 (0.01) & 1.07 (0.01) \\
\hline

\multirow[t]{3}{*}{IRAK4}
& Reference & 97.32 & 96.39 & 85.72 \\*
& TACFormer & 93.97 (0.11) & 92.49 (0.18) & 85.74 (0.09) \\*
& Unconditional & 3.71 (0.04) & 3.80 (0.04) & 3.74 (0.03) \\
\hline

\multirow[t]{3}{*}{EGFR}
& Reference & 90.99 & 84.94 & 76.33 \\*
& TACFormer & 90.86 (0.16) & 85.26 (0.23) & 76.40 (0.34) \\*
& Unconditional & 14.23 (0.06) & 13.41 (0.06) & 12.50 (0.04) \\
\hline

\multirow[t]{3}{*}{BCL2L1}
& Reference & 94.88 & 90.00 & 83.66 \\*
& TACFormer & 91.71 (0.38) & 88.58 (0.54) & 82.54 (0.67) \\*
& Unconditional & 2.51 (0.03) & 2.56 (0.03) & 2.58 (0.02) \\
\hline

\multirow[t]{3}{*}{SMARCA2}
& Reference & 95.63 & 87.08 & 81.61 \\*
& TACFormer & 79.30 (0.32) & 73.34 (0.22) & 67.61 (0.11) \\*
& Unconditional & 2.86 (0.03) & 2.75 (0.04) & 2.69 (0.04) \\
\hline

\multirow[t]{3}{*}{METTL14}
& Reference & \textbf{100.00} & \textbf{100.00} & \textbf{100.00} \\*
& TACFormer & \textbf{99.57 (0.14)} & \textbf{99.55 (0.15)} & \textbf{99.41 (0.13)} \\*
& Unconditional & 2.52 (0.03) & 2.89 (0.03) & 3.28 (0.02) \\
\hline

\multirow[t]{3}{*}{PTPN1}
& Reference & 90.67 & 86.33 & 84.33 \\*
& TACFormer & 81.92 (0.63) & 78.96 (0.65) & 77.22 (0.66) \\*
& Unconditional & 1.01 (0.02) & 1.00 (0.03) & 1.02 (0.03) \\
\hline

\multirow[t]{3}{*}{BCL2}
& Reference & 86.67 & 79.56 & 70.89 \\*
& TACFormer & 75.60 (0.17) & 71.92 (0.25) & 64.41 (0.48) \\*
& Unconditional & 1.42 (0.03) & 1.46 (0.03) & 1.37 (0.03) \\
\hline

\multirow[t]{3}{*}{SOS1}
& Reference & 90.40 & 69.20 & 42.40 \\*
& TACFormer & 86.51 (0.49) & 74.80 (0.27) & 51.12 (0.14) \\*
& Unconditional & 1.29 (0.01) & 1.36 (0.01) & 1.29 (0.01) \\
\hline

\multirow[t]{3}{*}{BCL6}
& Reference & \textbf{100.00} & \textbf{100.00} & 98.19 \\*
& TACFormer & 98.60 (0.09) & 98.52 (0.13) & 98.04 (0.14) \\*
& Unconditional & 1.75 (0.02) & 1.86 (0.02) & 2.03 (0.02) \\
\hline

\multirow[t]{3}{*}{MAPT}
& Reference & 90.86 & 86.86 & 84.00 \\*
& TACFormer & 89.00 (0.54) & 86.40 (0.54) & 83.60 (0.51) \\*
& Unconditional & 2.25 (0.02) & 2.06 (0.02) & 1.98 (0.01) \\
\hline

\multirow[t]{3}{*}{\textbf{Macro}}
& \textbf{Reference} & \textbf{92.27} & \textbf{86.72} & \textbf{78.33} \\*
& \textbf{TACFormer} & \textbf{84.73 (0.11)} & \textbf{80.60 (0.11)} & \textbf{74.01 (0.14)} \\*
& \textbf{Unconditional} & \textbf{5.02 (0.00)} & \textbf{5.02 (0.00)} & \textbf{5.02 (0.00)} \\
\hline

\end{longtable}
\endgroup

\paragraph{Zero-shot POIs.}

Table~\ref{tab:zero_shot_reference_generated_purity} compares held-out reference and generated $k$-NN purity for the four zero-shot POIs. At $k=5$, macro reference purity reaches $99.25\%$, compared with $30.52\%$ for generated molecules. The gap persists at $k=10$ and $k=20$. Generated 5-NN purity ranges from $56.23\%$ for WDR5 to $14.57\%$ for MAP3K13. Thus, the reference data show strong local POI organization, but the conditional model achieves much weaker enrichment under unseen protein conditions.

\begin{table}[t]
\caption{
\textbf{Reference versus conditional $k$-NN purity for four zero-shot POIs.}
Reference values are fixed across seeds. Generated values are reported as mean (sample standard deviation) across three generation seeds. All values are percentages. Macro values are unweighted means across the four POIs.
}
\label{tab:unseen_real_gen_knn}
\label{tab:zero_shot_reference_generated_purity}
\centering
\scriptsize
\setlength{\tabcolsep}{5pt}
\begin{tabular}{lcccccc}
\hline
\textbf{POI} &
\textbf{Ref. 5-NN} &
\textbf{Gen. 5-NN} &
\textbf{Ref. 10-NN} &
\textbf{Gen. 10-NN} &
\textbf{Ref. 20-NN} &
\textbf{Gen. 20-NN} \\
\hline
WDR5 &
\textbf{100.00} & \textbf{56.23 (0.10)} &
97.88 & \textbf{55.35 (0.35)} &
94.09 & \textbf{50.58 (0.44)} \\
DHFR &
98.60 & 31.05 (0.32) &
94.42 & 37.06 (0.28) &
90.70 & 45.18 (0.14) \\
FGFR3 &
\textbf{100.00} & 20.26 (0.15) &
\textbf{100.00} & 20.51 (0.07) &
\textbf{100.00} & 20.70 (0.05) \\
MAP3K13 &
98.40 & 14.57 (0.46) &
96.80 & 13.27 (0.31) &
95.30 & 12.71 (0.12) \\
\hline
\textbf{Macro} &
\textbf{99.25} &
\textbf{30.52 (0.06)} &
\textbf{97.27} &
\textbf{31.55 (0.11)} &
\textbf{95.02} &
\textbf{32.29 (0.14)} \\
\hline
\end{tabular}
\end{table}

\subsection{Chemical space organization of generated PROTACs}
\label{app:generated_chemical_space}

\begin{figure*}[!htbp]
\centering
\includegraphics[width=\textwidth]{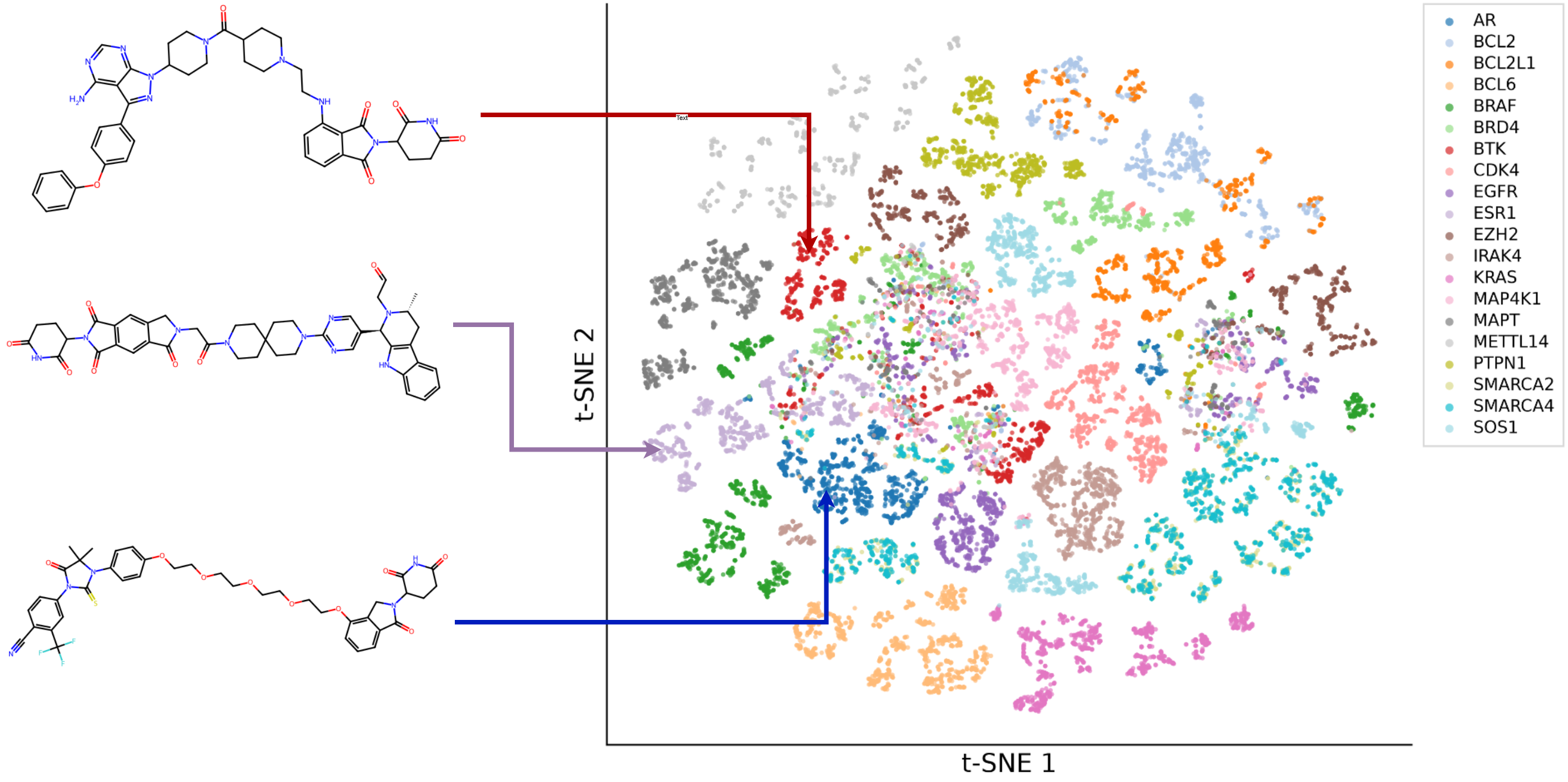}
\caption{
\textbf{Generated PROTACs retain the heterogeneous chemical-space organization observed in reference data.}
Two-dimensional t-SNE projection of novel molecules generated under the seen POI conditions, colored by conditioning POI. Molecules are represented using 2048-bit Morgan fingerprints with radius 2 and projected using a single common t-SNE fit. Example structures are novel generated PROTACs, with arrows indicating their conditioning POIs. 
}
\label{fig:generated_chemical_space}
\end{figure*}

Figure~\ref{fig:generated_chemical_space} visualizes molecules generated under the seen POI conditions. Each molecule is represented by a 2048-bit Morgan fingerprint with radius 2, and all generated molecules are projected jointly into a two-dimensional t-SNE space.

The embedding reveals POI-dependent regional structure, with molecules generated under the same condition can occupy multiple regions rather than a single compact cluster. This resembles the heterogeneous, multimodal pattern observed in the reference PROTAC data, where individual POIs also span distinct regions of chemical space. The visualization complements the quantitative neighborhood-purity results in Section~\ref{sec:seen_poi_results}.

\section{Generation dynamics during fine-tuning}

\subsection{Generation quality and conditional fidelity during fine-tuning}
\label{app:training_dynamics}

We monitor molecular generation quality throughout fine-tuning using validity, uniqueness, novelty, and internal diversity. These metrics capture different aspects of generation and do not necessarily improve together.

For the unconditional model, validity generally increases during fine-tuning, while novelty decreases. Uniqueness remains high in the monitoring samples, and internal diversity changes little across epochs. 

The conditional model shows a similar trade-off. Validity increases from $90.42\%$ at epoch 2 to $98.84\%$ at epoch 60. In contrast, uniqueness decreases from $97.64\%$ to $71.30\%$, while novelty decreases from $97.60\%$ to $72.33\%$. Internal diversity remains relatively stable, ranging from $0.775$ to $0.798$.

Conditional fidelity also improves throughout training. $\Delta_{\mathrm{NLL}}$ increases from $0.0155$ to $0.4155$, while generated 5-NN purity rises from $9.48\%$ to $87.12\%$, reaching its highest value of $89.72\%$ at epoch 58. These trends indicate that fine-tuning strengthens the model's preference for the correct POI condition and increases enrichment in target-associated chemical neighborhoods.

\begin{figure*}[!t]
\centering
\includegraphics[width=\textwidth]{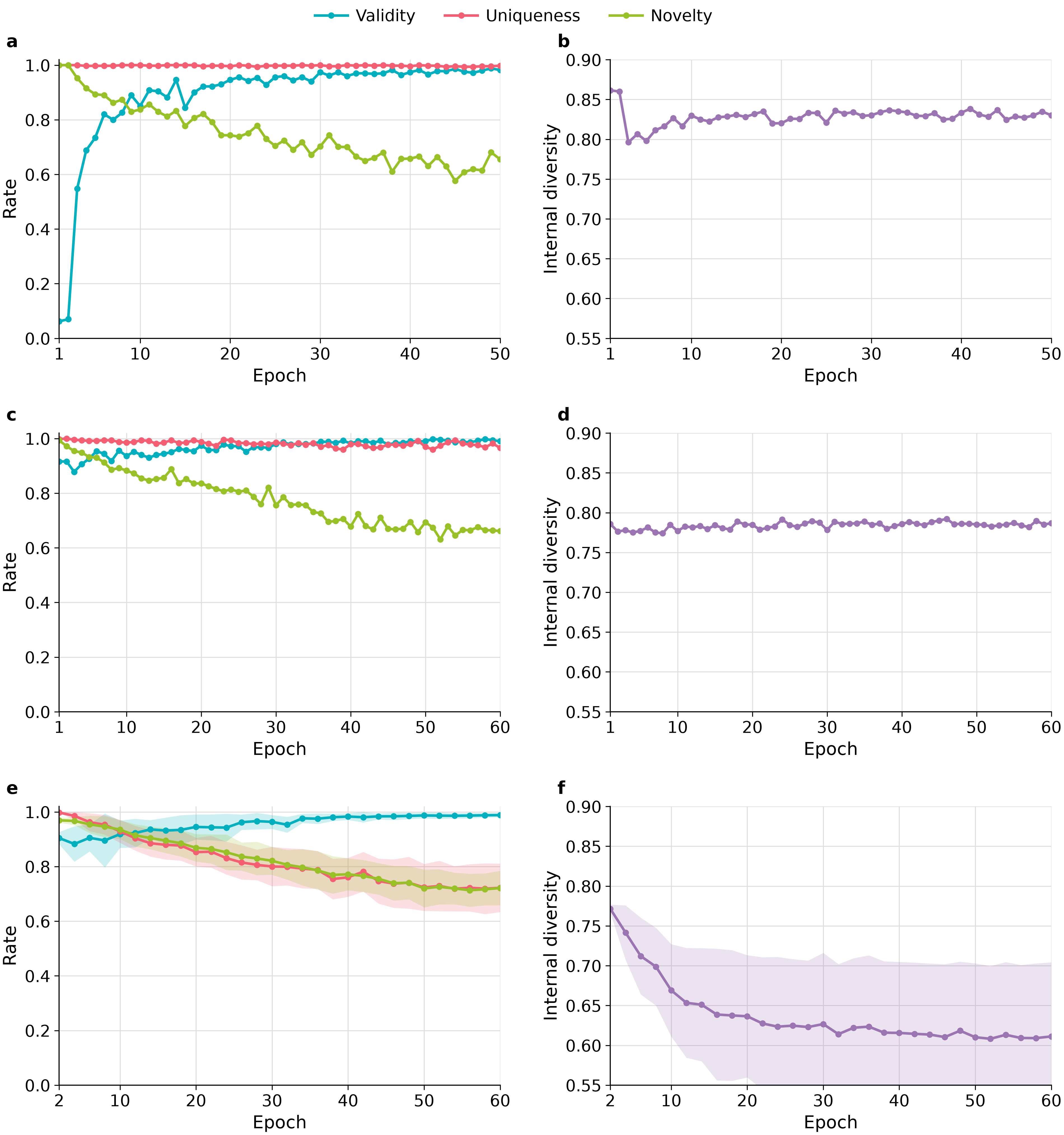}
\caption{
\textbf{Generation quality during molecular pretraining, unconditional fine-tuning, and conditional fine-tuning.}
Panels \textbf{(a,b)} show the molecular prior, \textbf{(c,d)} the unconditional model, and \textbf{(e,f)} the conditional model.
\textbf{(a,c,e)} Validity, uniqueness among valid molecules, and novelty among unique valid molecules across training epochs.
\textbf{(b,d,f)} Internal diversity among unique valid molecules.
For the conditional model, solid lines represent the mean across the seen POIs, and shaded regions indicate $\pm1$ standard deviation across targets.
}
\label{fig:training_dynamics}
\end{figure*}

\begin{figure*}[!t]
\centering
\includegraphics[width=\textwidth]{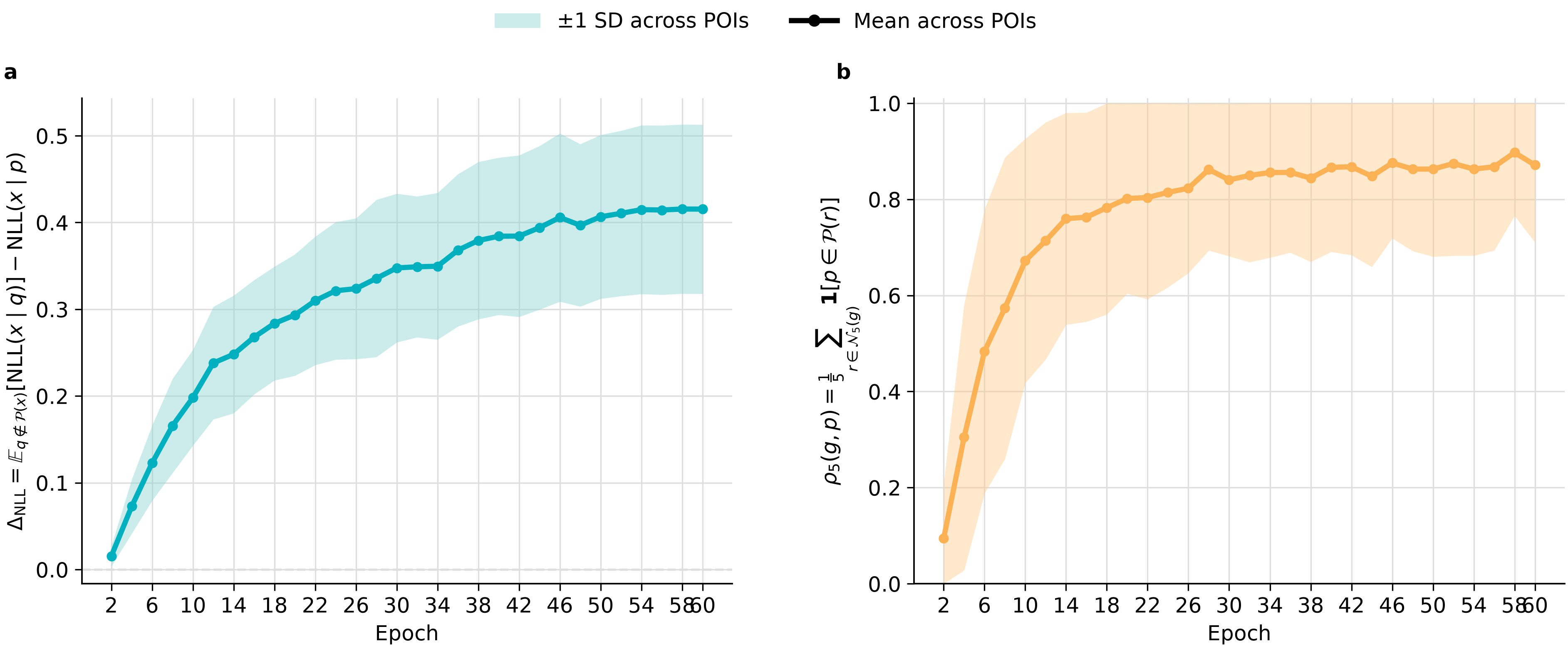}
\caption{
\textbf{Emergence of POI-specific conditioning during fine-tuning.}
\textbf{(a)} $\Delta_{\mathrm{NLL}}$ and \textbf{(b)} 5-NN purity across training epochs. Solid lines show the mean across the seen POIs, and shaded regions indicate $\pm1$ standard deviation. Likelihood evaluation uses a fixed set of validation molecules, while nearest-neighbor purity uses a fixed validation reference pool. At each monitored epoch, 500 molecules are generated per POI, with up to 25 unique valid molecules per target selected for 5-NN evaluation.
}
\label{fig:conditioning_dynamics}
\end{figure*}

Figures~\ref{fig:training_dynamics} and ~\ref{fig:conditioning_dynamics} show the generation and conditional-fidelity trajectories during fine-tuning.

\subsection{Checkpoint selection}
\label{app:checkpoint_selection}

\paragraph{Conditional model.}
Checkpoint selection for the conditional model must account for both generation quality and sensitivity to the protein condition. We therefore use a composite score that combines molecular generation quality with conditional fidelity. Checkpoint selection uses the validation set to compute
$\Delta_{\mathrm{NLL}}$ and generated 5-NN purity.

Generation quality is defined as
\begin{equation}
Q_{\mathrm{gen}}(e)
=
\left(
V_e U_e N_e D_e
\right)^{1/4},
\end{equation}
where $V_e$, $U_e$, $N_e$, and $D_e$ denote validity, uniqueness, novelty, and internal diversity at epoch $e$, respectively.

Conditional fidelity combines generated 5-NN purity with normalized likelihood preference:
\begin{equation}
F_{\mathrm{cond}}(e)
=
\left(
P_5(e)\,
\widetilde{\Delta}_{\mathrm{NLL}}(e)
\right)^{1/2},
\end{equation}
where
\begin{equation}
\widetilde{\Delta}_{\mathrm{NLL}}(e)
=
\frac{
\max(0,\Delta_{\mathrm{NLL}}(e))
}{
\max_{e'}\max(0,\Delta_{\mathrm{NLL}}(e'))
}.
\end{equation}

The final checkpoint score is
\begin{equation}
S_{\mathrm{ckpt}}(e)
=
100
\left(
Q_{\mathrm{gen}}(e)
F_{\mathrm{cond}}(e)
\right)^{1/2}.
\end{equation}

All generation-quality metrics and $P_5$ are expressed as fractions between 0 and 1. The normalization scales the positive likelihood preference relative to its maximum across monitored epochs. The score therefore rewards checkpoints that maintain molecular quality while producing stronger POI-associated chemical enrichment.

Table~\ref{tab:conditional_checkpoint_selection} reports the five highest-scoring checkpoints. Epoch 58 achieves the highest score ($86.82$), with $98.78\%$ validity, $71.04\%$ uniqueness, $71.77\%$ novelty, and $0.796$ internal diversity. It also achieves the highest monitored likelihood preference ($+0.4155$) and a generated 5-NN purity of $89.72\%$.

Epoch 46 achieves higher generation quality but lower conditional fidelity. Epoch 60 achieves slightly higher validity but lower 5-NN purity. Epoch 58 therefore provides the best balance between generation quality and POI-specific chemical steering under the defined score. We select checkpoint epoch 58 for the final conditional generation experiments.

\begin{table*}[t]
\caption{
\textbf{Conditional checkpoint selection during fine-tuning.}
The five highest-scoring checkpoints are reported. Generation quality combines validity, uniqueness, novelty, and internal diversity. Conditional fidelity combines generated 5-NN purity and normalized $\Delta_{\mathrm{NLL}}$. The final score balances both components. Bold values indicate the highest values among the reported checkpoints.
}
\label{tab:conditional_checkpoint_selection}
\centering
\scriptsize
\setlength{\tabcolsep}{5pt}
\begin{tabular}{lccccccccc}
\hline
\textbf{Epoch} &
\textbf{Validity (\%)} &
\textbf{Uniq. (\%)} &
\textbf{Novelty (\%)} &
\textbf{Diversity} &
\textbf{$\Delta_{\mathrm{NLL}}$} &
\textbf{5-NN (\%)} &
\textbf{$Q_{\mathrm{gen}}$} &
\textbf{$F_{\mathrm{cond}}$} &
\textbf{Score (\%)} \\
\hline
46
& 98.40 & \textbf{72.90} & \textbf{74.13}
& 0.795 & 0.4058 & 87.56
& \textbf{0.8063} & 0.9247 & 86.34 \\

60
& \textbf{98.84} & 71.30 & 72.33
& \textbf{0.796} & 0.4154 & 87.12
& 0.7981 & 0.9332 & 86.30 \\

52
& 98.63 & 72.03 & 72.85
& 0.795 & 0.4106 & 87.44
& 0.8009 & 0.9295 & 86.28 \\

56
& 98.66 & 71.34 & 71.38
& \textbf{0.796} & 0.4144 & 86.76
& 0.7952 & 0.9302 & 86.00 \\

\textbf{58}
& 98.78 & 71.04 & 71.77
& \textbf{0.796} & \textbf{0.4155}
& \textbf{89.72}
& 0.7957 & \textbf{0.9472}
& \textbf{86.82} \\
\hline
\end{tabular}
\end{table*}

\paragraph{Unconditional model.}
The unconditional model does not use protein conditions, so conditional-fidelity metrics are not applicable. We instead prioritize molecular validity while maintaining uniqueness, novelty, and internal diversity.

Epochs 51 and 58 achieve the highest observed validity ($99.80\%$). Compared with epoch 51, epoch 58 achieves higher uniqueness ($96.79\%$ versus $95.99\%$) and internal diversity ($0.790$ versus $0.785$), while maintaining comparable novelty ($66.46\%$ versus $67.43\%$). We therefore select epoch 58 for unconditional generation. Table~\ref{tab:unconditional_checkpoint_selection} reports the corresponding metrics.

\begin{table}[H]
\caption{
\textbf{Unconditional checkpoint selection during fine-tuning.}
The two checkpoints with the highest observed validity are compared. Metrics are computed from 500 generated molecules per epoch. Bold values indicate the highest values in each column.
}
\label{tab:unconditional_checkpoint_selection}
\centering
\scriptsize
\setlength{\tabcolsep}{5pt}
\begin{tabular}{lcccc}
\hline
\textbf{Epoch} &
\textbf{Validity (\%)} &
\textbf{Uniq. (\%)} &
\textbf{Novelty (\%)} &
\textbf{Diversity} \\
\hline
51
& \textbf{99.80}
& 95.99
& \textbf{67.43}
& 0.785 \\

\textbf{58}
& \textbf{99.80}
& \textbf{96.79}
& 66.46
& \textbf{0.790} \\
\hline
\end{tabular}
\end{table}

\end{document}